\documentclass{article} % For LaTeX2e
\usepackage{preprint,times}

\usepackage{amsmath,amsfonts,bm}

\def\eqref#1{equation~\ref{#1}}
\def\1{\bm{1}}

\DeclareMathAlphabet{\mathsfit}{\encodingdefault}{\sfdefault}{m}{sl}
\SetMathAlphabet{\mathsfit}{bold}{\encodingdefault}{\sfdefault}{bx}{n}

\usepackage{hyperref}
\hypersetup{hidelinks}
\usepackage{url}

\usepackage{booktabs}
\usepackage{colortbl}
\definecolor{postshade}{gray}{0.94}
\usepackage{graphicx}
\setkeys{Gin}{pagebox=cropbox,clip=true}
\usepackage{cleveref}
\usepackage{placeins}
\title{Latent Watermarks under\\Generative Editing: A Benchmark\\and Analysis of Detection Survival}

\author{Sung Ju Lee, Nam Ik Cho 
\\
Department of ECE \& INMC \\
Seoul National University \\
Seoul, South Korea \\
\texttt{\{thomas11809,nicho\}@snu.ac.kr} \\
}

\begin{document}

\maketitle

\begin{abstract}
Ordinary prompt-based editing can cause latent watermark detection to fail without explicitly targeting the watermark. We benchmark eight watermark methods against five editors across four generative backbones, four editing strengths, and five semantic categories, with edit-validity and threshold checks. Separating editing from seven subsequent distortions reveals that editing alone primarily distinguishes Tree-Ring, while added distortions expose a broader spectrum of detection survival. Sequential edits reveal a second hidden difference: score separation can decline while detection rates remain near their ceiling. Across methods, standardized clean score separation ($d'$) organizes composite-survival tiers, whereas spatial overlap adds little to predicting edit-only survival beyond clean detectability. Embedding-strength interventions in two methods link higher clean separation to higher post-edit separation. In HSTR, the margin contrast is positive, while the angular layout contrast at matched clean separation remains unresolved. Together, outcome decomposition and continuous separation expose differences hidden by aggregate TPR. Method tiers are stable under threshold recalibration at the main operating points and alternative composite weights. Clean $d'$ is thus a useful empirical diagnostic within this benchmark, with mixed transfer to unseen methods. Code and supporting artifacts are planned for a separate release.
\end{abstract}

\begin{figure}[!ht]
\centering
\includegraphics[width=\linewidth]{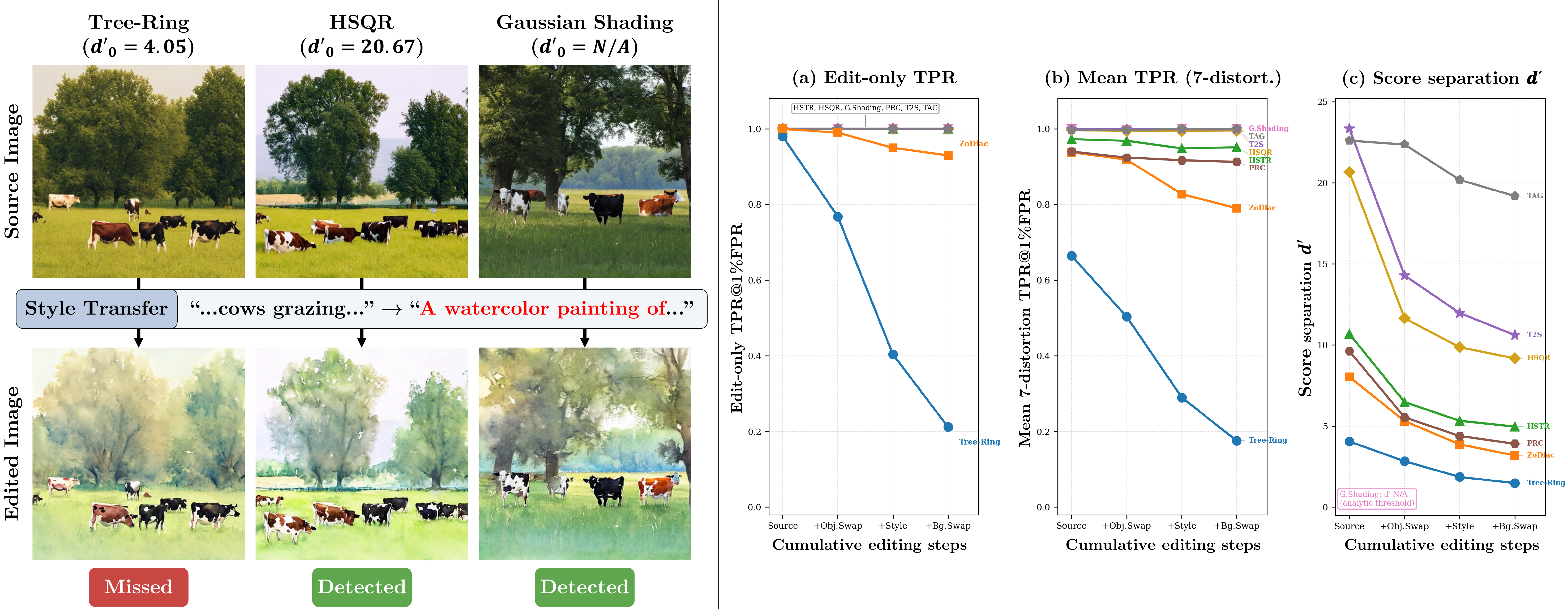}
\caption{\textbf{Watermark survival under generative editing.} Left: InfEdit Style Transfer examples; $d'_0$ summarizes each method's source bank. Right: a separate cumulative InfEdit cascade (Source, Object Swap, Style Transfer, Background Swap), showing TPR before further distortion, mean TPR over seven separately applied distortions, and score separation. The TPR plots include all eight methods; $d'$ is available for seven, with GS shown as N/A. Source counts and the five-step extension are reported in \Cref{sec:multistep}.}
\label{fig:teaser}
\end{figure}

\section{Introduction}
\label{sec:intro}

Image watermarking provides a persistent signal that can complement provenance records such as C2PA~\citep{c2pa2026spec}. Generative approaches modify the decoder~\citep{fernandez2023stable} or initial sampling noise $z_T$~\citep{wen2024tree,yang2024gaussian}. Post-hoc approaches also embed messages in VAE-encoded image latents $z_0$, using learned perturbations~\citep{bui2023rosteals}, frequency-domain optimization~\citep{guo2024freqmark}, or direct phase modulation~\citep{lee2026phasemark}.

Text-guided editors can resynthesize image content without explicitly targeting its watermark~\citep{brooks2023instruct,kulikov2025flowedit}. This has motivated editing-aware watermark designs, including Robust-Wide~\citep{hu2024robustwide}, JIGMARK~\citep{pan2024jigmark}, and VINE~\citep{lu2024robust}. Their development makes systematic evaluation across watermark designs and editing regimes increasingly important.

Existing evaluations span distortion and regeneration stress tests~\citep{an2024waves}, broad editing benchmarks~\citep{lu2024robust}, and semantic perturbations of in-processing watermarks~\citep{nakra2026semantic}. SafeMark additionally trains editors to preserve watermarks and evaluates subsequent distortions~\citep{wu2026safemark}. A prior study of ordinary prompt-based editing proposed Semantic Imprinting to interpret differences in survival~\citep{lee2026hypothesis}. Building on these studies, we ask: under a common protocol, how much of the robustness gap arises from editing versus subsequent distortions, and which measured properties consistently organize that gap?

We evaluate eight initial-noise or inversion-based latent watermark designs under five editors, four generative backbones, four strengths, and five semantic categories, with validity and threshold checks. Editing alone primarily separates Tree-Ring from the other methods; seven subsequent distortions reveal a broader spectrum. \Cref{fig:teaser} illustrates these differences through examples and a cumulative editing sequence. Across the evaluated methods, clean score separation, $d'$, organizes composite survival, while spatial overlap adds little when predicting edit-only survival. Within-method interventions and held-out analyses test the usefulness and limits of this empirical diagnostic.

Our contributions are \textbf{(1) a controlled benchmark} with explicit operating points and validity checks; \textbf{(2) an outcome decomposition} separating edit-only survival from subsequent-distortion robustness; and \textbf{(3) an empirical diagnosis} showing that clean score separation organizes composite survival and tracks within-method changes, while spatial tests and held-out evaluation delimit its explanatory and predictive scope.
\section{Related Work}
\label{sec:related}

Watermarks can be integrated through decoder fine-tuning~\citep{fernandez2023stable}, intermediate diffusion states~\citep{huang2024robin}, or initial-noise encoding. Post-hoc VAE-latent methods include RoSteALS~\citep{bui2023rosteals}, FreqMark~\citep{guo2024freqmark}, and PhaseMark~\citep{lee2026phasemark}; their image latents $z_0$ differ from sampling noise $z_T$. Our evaluation focuses on eight initial-noise or inversion-based designs. Tree-Ring~\citep{wen2024tree} and HSTR/HSQR~\citep{lee2025semantic} use Fourier patterning, whereas ZoDiac~\citep{zhang2025attack} fits a latent post hoc. PRC~\citep{gunn2025undetectable} uses a pseudorandom error-correcting code and T2SMark~\citep{yang2026t2smark}, abbreviated T2S, uses tail-truncated sampling; Gaussian Shading~\citep{yang2024gaussian} and TAG-WM~\citep{chen2025tag}, abbreviated GS and TAG, use cryptographic bit encoding. SEAL binds watermark verification to image semantics to address forgery~\citep{arabi2025seal}; our outcome is signal survival, distinct from semantic authenticity.

\noindent\textbf{Prompt-based generative editing.}
Prompt-to-Prompt (P2P)~\citep{hertz2022prompt} edits images via cross-attention injection; Null-text Inversion~\citep{mokady2023null} extends this to real images (P2P+NTI). InfEdit~\citep{xu2024inversion} performs inversion-free, attention-controlled editing on a consistency model. FlowEdit~\citep{kulikov2025flowedit} edits flow models such as SD3 and FLUX via source-target velocity-field differences, without inversion or attention manipulation. These paradigms motivate our editor coverage.

\noindent\textbf{Watermarking under editing.}
Robust-Wide incorporates instruction-driven editing into watermark training~\citep{hu2024robustwide}; JIGMARK learns from edited image pairs without backpropagating through the editor~\citep{pan2024jigmark}; VINE uses generative priors and surrogate degradations~\citep{lu2024robust}. SafeMark instead optimizes the editor to preserve watermark decoding~\citep{wu2026safemark}. We evaluate ordinary editors without such watermark-preservation objectives.

\noindent\textbf{Benchmarks and empirical diagnostics.}
WAVES evaluates distortions, regeneration, and adversarial attacks~\citep{an2024waves}, while W-Bench covers global/local editing, regeneration, and image-to-video generation~\citep{lu2024robust}. \citet{nakra2026semantic} study semantic perturbations of Stable Signature, Tree-Ring, and GS; \citet{zhao2024invisible} show that generative regeneration provably removes watermarks whose image perturbation is $\ell_2$-bounded. The Semantic Imprinting hypothesis~\citep{lee2026hypothesis} offers a candidate interpretation of survival differences. We extend this empirical question through a shared multi-editor protocol, separate editing and subsequent-distortion outcomes, and examine spatial overlap and clean separation through complementary diagnostics, within-method controls, and held-out analyses.

\section{Benchmark and Analysis Framework}
\label{sec:framework}

\begin{figure}[t]
\centering
\includegraphics[width=\linewidth]{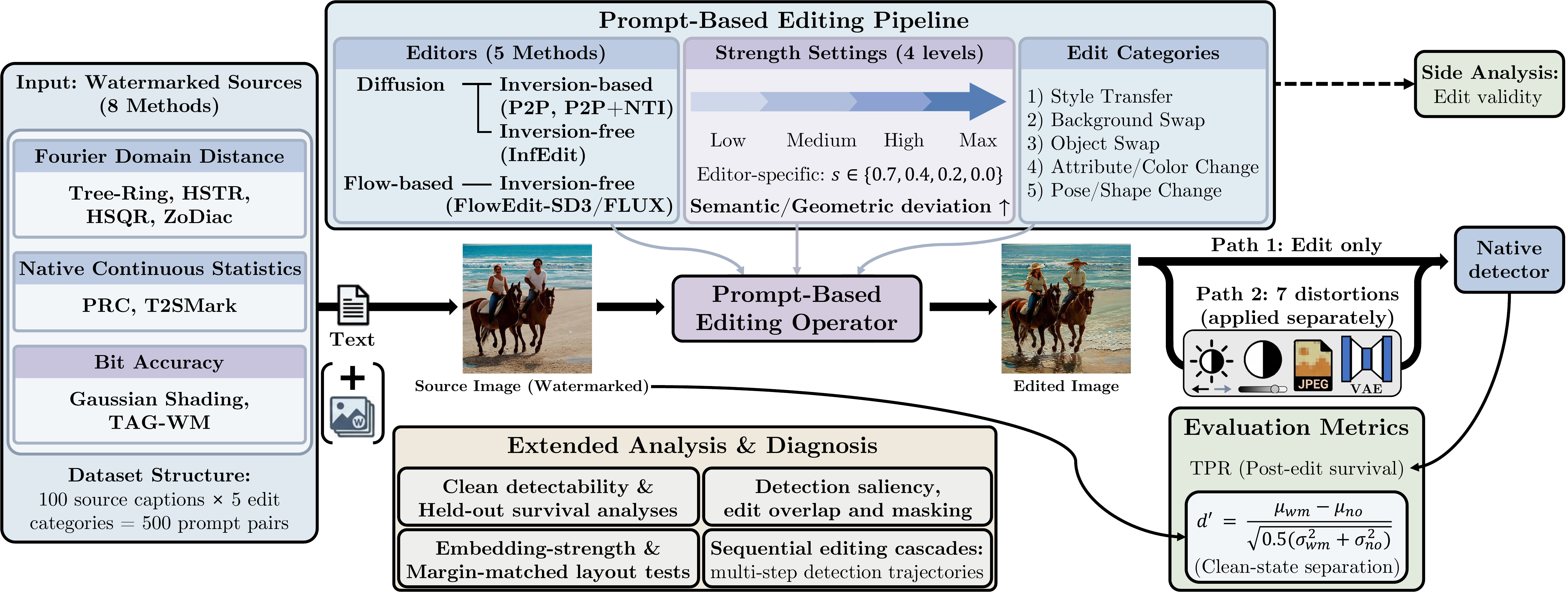}
\caption{Benchmark and analysis framework. Eight watermark methods are evaluated under five editors, four strength settings, and five edit categories. Detection is measured before and after seven separately applied distortions. Low/Medium/High/Max and the deviation arrow indicate the intended progression of each editor's control settings; Max is the strongest setting tested.}
\label{fig:framework_overview}
\end{figure}

\Cref{fig:framework_overview} summarizes the benchmark and analysis framework.

\subsection{Threat Model and Benchmark Matrix}
\label{sec:benchmark}
\label{sec:threat_model}

A watermark is introduced at the source image stage, either in the initial sampling latent at generation time or by fitting a post-hoc inverted latent to an existing image. The published image then undergoes ordinary prompt-based generative editing, without optimization against the watermark. Detection is blind: the detector operates on the edited output without the original image. Our operational outcome is watermark detection survival; edit validity, perceptual quality, and cryptographic security are separate evaluation dimensions.

Eight watermark methods and five editors spanning four generative backbones are evaluated at four settings, yielding $8\times5\times4=160$ method--editor--strength cells. Each cell uses 100 MS-COCO~\citep{lin2014microsoft} source captions, with one target edit per caption in each of five categories. These repeated edits share the same source-caption identities.

\paragraph{Separating editing from subsequent distortions.}
\emph{Edit only} measures TPR immediately after editing. \emph{Post-distortion mean} averages TPR over seven distortions applied separately to that edited output. Avg$_5$ averages either measure over the five categories. The \emph{8-condition composite} includes Edit only and all seven distortion TPRs, averaged within category and then across categories. The main table and strength-sweep figure separate the first two outcomes; the appendix also tabulates the composite sweep used in the clean-$d'$ analyses. The seven distortions are brightness, contrast, JPEG, Gaussian blur, BM3D, and two learned re-encodings: a fixed stress suite (\Cref{sec:app_hyperparams}).

Thresholds target a nominal $1\%$ false-positive rate (FPR); \Cref{sec:app_hyperparams,sec:app_fpr} specify the operating rules and measured FPRs.

\subsection{Evaluation Metrics and Edit Validity}
\label{sec:metrics}

\paragraph{Detection survival and score regimes.} All methods are evaluated through their native statistic and a declared decision rule. Six methods use continuous statistics, while TAG uses bit accuracy; these seven use strict score $>$ empirical null $q_{99}$. GS uses the analytic bit-accuracy rule $\ge147/256$, and TPR reports the fraction of surviving detections under each rule. Continuous-separation analyses use the six informative native statistics; TAG's finite $d'$ is limited by clean-score saturation, and GS has no compatible empirical null estimate in this benchmark.

\paragraph{Standardized score separation.}
\label{sec:dprime_def}
The sensitivity index $d'$ from signal detection theory~\citep{green1966signal,macmillan2005detection} compares the watermarked and null score distributions. Let $\mu_{\mathrm{wm}},\sigma_{\mathrm{wm}}^2$ and $\mu_{\mathrm{no}},\sigma_{\mathrm{no}}^2$ denote their respective means and variances:
\begin{equation}
d' = \frac{\mu_{\mathrm{wm}} - \mu_{\mathrm{no}}}{\sqrt{0.5\,(\sigma_{\mathrm{wm}}^2 + \sigma_{\mathrm{no}}^2)}},
\label{eq:dprime}
\end{equation}
with scores oriented so that larger values indicate stronger watermark evidence. \emph{Clean $d'$}, also called clean detection margin, uses pre-edit source scores. Post-edit $d'$ applies the same definition to edited scores with the corresponding null reference. This standardized separation is distinct from distance to an operating threshold. \Cref{sec:app_score_functions} gives score conventions; the controlled analyses specify their aggregation units.

\paragraph{Edit validity and severity.}
\label{sec:severity_metric}
Relative to the watermarked editing input, LPIPS~\citep{zhang2018unreasonable} measures perceptual distance and DINOv2~\citep{oquab2023dinov2} measures structural/content similarity. $\Delta$CLIP~\citep{radford2021learning} and $\Delta$BLIP2 ITM~\citep{li2023blip2} measure changes in target-text alignment, using cosine similarity and positive matching probability, respectively. LPIPS also serves as a per-image edit-magnitude covariate in the severity analysis. \Cref{sec:app_validity_metrics,sec:app_severity} detail these metrics and their use.

\subsection{Candidate Explanations and Controlled Tests}
\label{sec:saliency}

We examine two candidate explanations for survival differences: the spatial relationship between detector sensitivity and editing changes, and clean score separation. We then vary embedding strength within HSTR and Tree-Ring, vary HSTR angular layout at matched clean separation, and test whether the observed associations transfer to held-out methods after accounting for edit magnitude.

\paragraph{Detection saliency.} For a detector score $S(x)$ evaluated on the RGB query $x$, saliency measures image-space sensitivity:
\begin{equation}
M_{\mathrm{sal}}(x) = \mathrm{Norm}\!\left(\frac{1}{C}\sum_{c=1}^{C}\left|\frac{\partial S(x)}{\partial x_c}\right|\right),
\label{eq:saliency}
\end{equation}
where $C$ is the channel count and $\mathrm{Norm}$ is min-max normalization to $[0,1]$. \Cref{sec:app_score_functions} describes the gradient chains and spatial surrogates.

\paragraph{Maps of editing changes.} Cross-attention differences summarize source-to-target attention changes during editing; pixel changes measure source-to-edited RGB differences. Qualitative comparisons use attention differences for P2P, P2P+NTI, and InfEdit, and pixel changes for FlowEdit. Quantitative overlap uses pixel changes for every editor, providing a common image-space comparison. \Cref{sec:app_edit_region_eqs} defines the maps.

\section{Robustness under Generative Editing}
\label{sec:results}

\subsection{Editing-Only Survival and Subsequent Distortions}
\label{sec:benchmark_results}

\Cref{tab:main_verification} reports Avg$_5$ TPR for all eight methods and five editors at the main operating point, separating the edited output before additional distortion (Edit only) from the mean over seven distortions applied separately after editing (Post-distortion mean). Editing alone already distinguishes Tree-Ring: its Edit-only TPR ranges from $0.632$ to $0.804$, whereas every other method lies between $0.986$ and $1.000$. The smallest within-editor gap is $0.196$. Tree-Ring's lowest Edit-only TPR ($0.632$) occurs under P2P, which shares the SD2.1 source backbone (\Cref{sec:app_hyperparams}).

% Source: audit_reports-astra/_work/ec7_decomposition.csv; main strengths only.
\begin{table}[t]
\centering
\caption{Watermark survival at the main editing operating points. Entries are Avg$_5$ TPR over five edit categories. Edit only precedes added distortion; shaded Post-dist. columns average seven distortions applied separately to that output. P2P, P2P+NTI, InfEdit, and FlowEdit-SD3 use $s{=}0.4$; FlowEdit-FLUX uses $s{=}0.2$, $n_{\max}{=}23$. \Cref{sec:app_hyperparams} lists native editor settings.}
\label{tab:main_verification}
\scriptsize
\setlength{\tabcolsep}{1.8pt}
\begin{tabular}{l*{5}{r>{\columncolor{postshade}}r}}
\toprule
& \multicolumn{2}{c}{P2P} & \multicolumn{2}{c}{P2P+NTI} & \multicolumn{2}{c}{InfEdit} & \multicolumn{2}{c}{FlowEdit-SD3} & \multicolumn{2}{c}{FlowEdit-FLUX} \\
\cmidrule(lr){2-3}\cmidrule(lr){4-5}\cmidrule(lr){6-7}\cmidrule(lr){8-9}\cmidrule(lr){10-11}
Method & Edit only & Post-dist. & Edit only & Post-dist. & Edit only & Post-dist. & Edit only & Post-dist. & Edit only & Post-dist. \\
\midrule
Tree-Ring & 0.632 & 0.356 & 0.804 & 0.471 & 0.682 & 0.444 & 0.692 & 0.403 & 0.734 & 0.427 \\
ZoDiac & 0.996 & 0.818 & 1.000 & 0.873 & 0.992 & 0.908 & 0.986 & 0.877 & 0.992 & 0.880 \\
PRC & 0.994 & 0.800 & 1.000 & 0.861 & 1.000 & 0.928 & 1.000 & 0.861 & 1.000 & 0.848 \\
HSTR & 1.000 & 0.934 & 1.000 & 0.952 & 1.000 & 0.965 & 1.000 & 0.944 & 1.000 & 0.937 \\
HSQR & 1.000 & 0.991 & 1.000 & 0.996 & 1.000 & 0.997 & 1.000 & 0.992 & 1.000 & 0.991 \\
TAG & 1.000 & 0.999 & 1.000 & 0.997 & 1.000 & 0.997 & 1.000 & 0.998 & 1.000 & 0.997 \\
T2S & 1.000 & 0.998 & 1.000 & 0.999 & 1.000 & 0.999 & 1.000 & 0.999 & 1.000 & 0.999 \\
GS & 1.000 & 0.998 & 1.000 & 0.999 & 1.000 & 0.999 & 1.000 & 0.999 & 1.000 & 0.998 \\
\bottomrule
\end{tabular}
\end{table}

Subsequent distortions widen the Tree-Ring gap and separate ZoDiac, PRC, and HSTR from the near-ceiling methods. This broader spectrum also appears in the 8-condition composite used for the clean-$d'$ analyses. \Cref{tab:app_main_legacy8} reports the composite's mean and worst-category values together with the clean $d'$ values used in those analyses.

\paragraph{Operating-point sensitivity.} Realized FPRs vary by method around the nominal $1\%$ (\Cref{sec:app_fpr}). Recalibrating on a 600-image null bank changes composite TPR by at most $0.042$ at the five main operating points and preserves every pairwise order among the seven empirically thresholded methods whose original gap is at least $0.02$. Its 500-image cross-fitted counterpart yields method-wise condition-averaged FPRs of $0.8$--$1.2\%$. Stricter rules also preserve the ordering of the five tested methods (\Cref{sec:app_zerofp,sec:app_recalib}); \Cref{sec:app_seedci} reports repeated-seed intervals.

\subsection{Sensitivity to Editor-Specific Settings}
\label{sec:strength}

\begin{figure}[t]
\centering
\includegraphics[width=\linewidth]{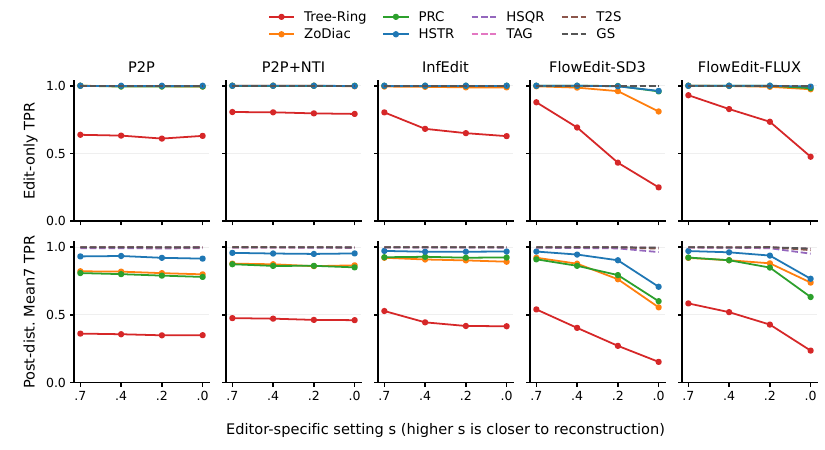}
\caption{Edit-only TPR (top) and the mean TPR over seven separately applied post-edit distortions (bottom), across eight methods, five editors and four settings. Each point averages five categories equally. Higher $s$ is closer to reconstruction within an editor. \Cref{tab:app_sweep} retains the corresponding 8-condition composite sweep.}
\label{fig:strength_sweep_main}
\end{figure}

\Cref{fig:strength_sweep_main} separates edit-only survival from post-distortion Mean7 across the full sweep. Tree-Ring remains below the other methods in edit-only survival; ZoDiac, PRC, and HSTR also lose survival at some settings, while HSQR, TAG, T2S, and GS remain at the ceiling. Subsequent distortions expose a broader spread. Both outcomes show greater setting dependence for the flow-based editors than for P2P or P2P+NTI. Setting definitions appear in \Cref{sec:app_hyperparams}.

Style Transfer is the most frequent worst category under the composite (77 of 160 cells, $48\%$); category details appear in \Cref{sec:app_catcond}.

\subsection{Edit Validity and Sanity Checks}
\label{sec:edit_validity}

\Cref{tab:edit_validity_main} reports perceptual and positive average target-alignment changes across the five original methods, with an HSQR example showing object changes. At the selected budget, FlowEdit-FLUX yields smaller changes: LPIPS and $\Delta$BLIP2 ITM are $0.187$ and $0.064$, versus $0.205$ and $0.321$ for FlowEdit-SD3. \Cref{sec:app_flux} gives the solver-budget sweep.

% Source: closeout_20260916/validity_t5_uniform256.csv and locked_T5_all_cells.csv
\begin{table}[t]
\centering
\caption{Edit magnitude and target alignment relative to the watermarked input at the main operating points. Top: each editor aggregates Tree-Ring, ZoDiac, HSTR, HSQR, and GS over five categories and 100 source captions. LPIPS is a distance, DINO a similarity, and the two deltas measure changes in target alignment; \Cref{sec:app_validity_metrics} defines the metrics. Bottom: a selected HSQR Object Swap example (source 59), shown for illustration rather than success-rate estimation.}
\label{tab:edit_validity_main}
\scriptsize
\begin{tabular}{lcccc}
\toprule
Editor & LPIPS & DINO & $\Delta$CLIP & $\Delta$BLIP2 ITM \\
\midrule
P2P           & 0.282 & 0.793 & 0.020 & 0.177 \\
P2P+NTI       & 0.283 & 0.747 & 0.025 & 0.202 \\
InfEdit       & 0.250 & 0.751 & 0.034 & 0.308 \\
FlowEdit-SD3  & 0.205 & 0.751 & 0.044 & 0.321 \\
FlowEdit-FLUX & 0.187 & 0.812 & 0.018 & 0.064 \\
\bottomrule
\end{tabular}

\par\vspace{6pt}
\includegraphics[width=\linewidth]{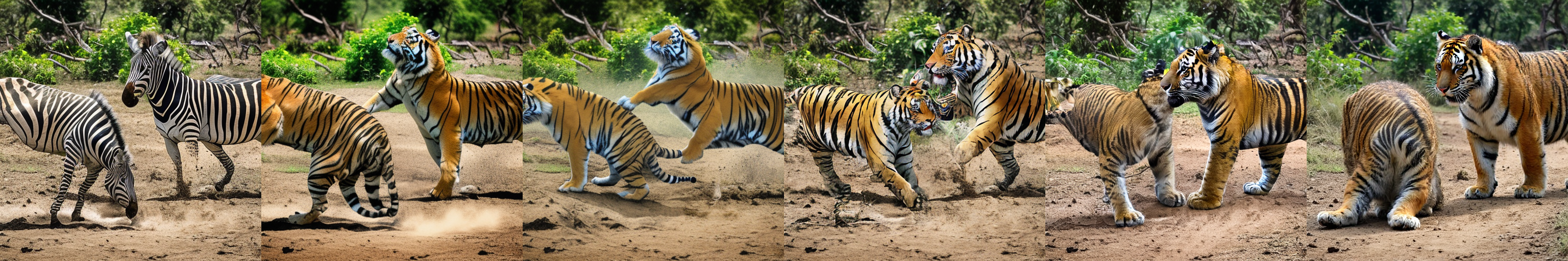}\\[2pt]
\makebox[0.166\linewidth]{\scriptsize HSQR source}%
\makebox[0.166\linewidth]{\scriptsize P2P}%
\makebox[0.166\linewidth]{\scriptsize P2P+NTI}%
\makebox[0.166\linewidth]{\scriptsize InfEdit}%
\makebox[0.166\linewidth]{\scriptsize FlowEdit-SD3}%
\makebox[0.166\linewidth]{\scriptsize FlowEdit-FLUX}
\par\vspace{3pt}\raggedright\footnotesize
Target: ``Two tigers in a jungle area fighting in a dirt area.'' All outputs use the same source and target prompt at the main editor-specific settings in \Cref{tab:main_verification}.
\end{table}

\section{What Explains the Robustness Gap?}
\label{sec:explain}

We first test whether spatial overlap adds predictive information for edit-only survival, then relate clean separation to composite survival across methods. Within-method interventions probe changes in score separation, and a final analysis tests severity adjustment and transfer of per-image detection predictions.

\subsection{Testing a Spatial Explanation}
\label{sec:spatial}

A spatial account suggests that editing preserves detection by avoiding detector-sensitive regions. We test source sensitivity, its ranking through masking, and overlap with editing changes.

\begin{figure}[t]
\centering
\includegraphics[width=\linewidth]{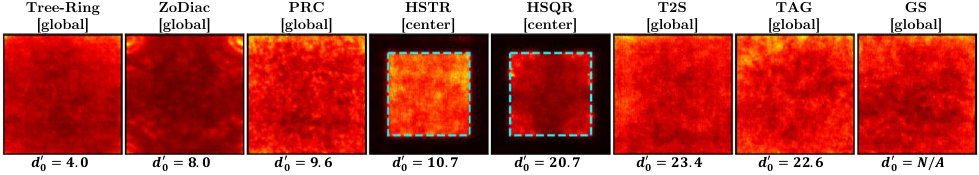}\\[-1pt]
\textbf{(a)}\;\small Source sensitivity: eight methods\par
\vspace{4pt}
\includegraphics[width=\linewidth]{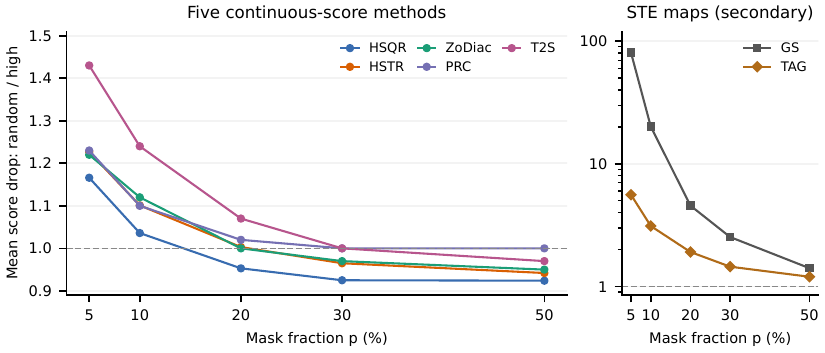}\\[-1pt]
\textbf{(b)}\;\small Masking: seven methods\par
\vspace{5pt}
\begin{minipage}[t]{0.42\linewidth}
\centering\footnotesize
\begin{tabular}{@{}lr@{}}
\toprule
\multicolumn{2}{c}{Mean top-10\% IoU (8 methods)} \\
\midrule
Observed & 0.022 \\
Uniform reference & 0.053 \\
\bottomrule
\end{tabular}
\end{minipage}\hfill
\begin{minipage}[t]{0.56\linewidth}
\centering\footnotesize
\begin{tabular}{@{}lr@{}}
\toprule
\multicolumn{2}{c}{Edit-only prediction (6 methods)} \\
\midrule
Pooled $\Delta$log-loss & $-0.0004$ \\
LOMO $\Delta$log-loss & $+0.0005$ \\
\bottomrule
\end{tabular}
\end{minipage}
\\[2pt]\textbf{(c)}\;\small Overlap with RGB editing changes\par
\caption{Spatial diagnostics. \textbf{(a)} Mean source sensitivity ($n{=}100$ per method); cyan guides mark the HSTR/HSQR latent-crop extent. Normalized maps do not compare absolute sensitivity. \textbf{(b)} Random-to-high-saliency mean score-drop ratios; $1$ means equal mean damage. GS/TAG use STE maps and a separate log axis. Low-saliency results remain in \Cref{fig:masking_ablation}. \textbf{(c)} The uniform IoU reference is geometric; $\Delta$log-loss adds overlap to clean $d'$ ($-$ is better).}
\label{fig:mechanism}
\end{figure}

\paragraph{Source-map distribution.}
\label{sec:spatial_disentangle}
\Cref{fig:mechanism}(a) shows central RGB sensitivity for HSTR/HSQR and broader maps for other methods. Spatial statistics show no monotone ordering with clean $d'$ or composite TPR (\Cref{tab:app_spatial}); these maps measure sensitivity rather than watermark location.

\paragraph{Masking as a ranking check.}
\label{sec:ablation}
We mask equally sized pixel sets in source images before editing. Random-to-high-saliency score-drop ratios lie on both sides of $1$ for continuous scores at larger fractions; GS/TAG remain above $1$ (\Cref{fig:mechanism}b). Low-saliency masking is not uniformly less damaging: HSTR's low-saliency drop exceeds its high-saliency drop at $p{=}30\%$ and $50\%$. The tested ranking thus offers no consistent targeted-removal advantage (\Cref{fig:masking_ablation,tab:app_ablation,tab:app_ablation_ext}).

\paragraph{Overlap and edit-only survival.}
\label{sec:ec2}
\Cref{fig:mechanism}(c) uses RGB changes for every editor. IoU is below the independent-uniform reference; adding overlap to clean $d'$ barely reduces pooled log-loss and slightly worsens leave-one-method-out (LOMO) prediction. Tree-Ring supplies most informative held-out variation. \Cref{sec:app_spatial_full,sec:app_mechanism_ext} give the quantitative and qualitative comparisons.

\FloatBarrier

\subsection{Clean Detectability Organizes Composite Survival}
\label{sec:margin}

\begin{figure}[t]
\centering
\begin{minipage}[t]{0.24\linewidth}
\centering
\includegraphics[width=\linewidth]{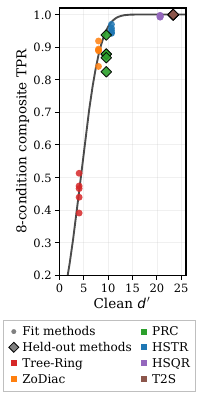}\\[2pt]
\textbf{(a)}
\end{minipage}\hspace{10pt}
\begin{minipage}[t]{0.24\linewidth}
\centering
\includegraphics[width=\linewidth]{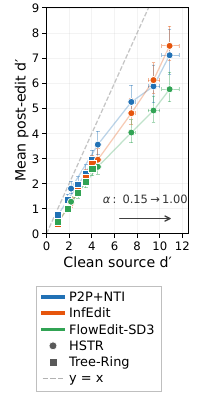}\\[2pt]
\textbf{(b)}
\end{minipage}\hspace{10pt}
\begin{minipage}[t]{0.4273\linewidth}
\centering
\includegraphics[width=\linewidth]{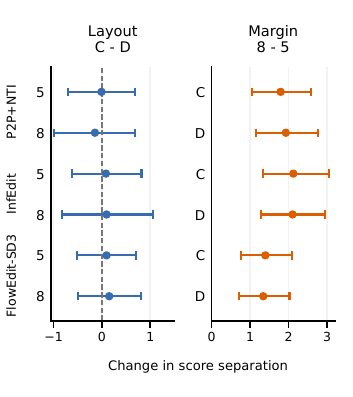}\\[2pt]
\textbf{(c)}
\end{minipage}
\caption{\textbf{(a)} Clean $d'$ versus 8-condition composite TPR with the global probit summary; circles denote fitted methods and diamonds held-out methods. \textbf{(b)} Strength intervention: clean versus mean category-level pre-distortion post-edit $d'$, with marginal 95\% bootstrap intervals. The arrow marks increasing $\alpha$ along each series. \textbf{(c)} HSTR effects on separation of source-mean scores, with paired-complete 95\% bootstrap intervals. Left rows give matched clean targets 5/8; right rows fix Compact (C) or Distributed (D). The two effect axes have different ranges. All six layout intervals include zero; all six margin intervals are positive. \Cref{sec:app_editor_aware,sec:app_ea6_detail,sec:app_eb3_effects} give the estimators.}
\label{fig:dprime_intervention}
\end{figure}

Edit-only survival is near ceiling for most methods, so we examine \emph{8-condition composite TPR}. Seven of its eight components include added distortions. Its tiers---Tree-Ring $<$ \{ZoDiac, PRC\} $<$ HSTR $<$ \{HSQR, TAG, T2S, GS\}---are unchanged in all 20 settings when any distortion is dropped, the Edit-only term is double-weighted, or only the seven distortions are averaged (\Cref{sec:app_main_legacy8}). Among methods with informative clean $d'$, these tiers follow its order. \Cref{fig:dprime_intervention}(a) shows six methods at the main operating points: each point averages five categories, sharing method-level clean $d'$ across editors. This aggregate association is distinct from per-image edit-only prediction.

\label{sec:probit}
A global probit curve fitted to four methods summarizes this association, with RMSE $0.086$ on 40 held-out PRC/T2S cells. \Cref{sec:app_editor_aware} distinguishes this global fit from condition-specific fits and their method-wise LOMO errors.

\paragraph{Sequential edits and score separation.}
Continuous separation can reveal changes hidden by ceiling TPR. \Cref{fig:teaser} shows the first three cumulative InfEdit edits, and \Cref{sec:multistep} extends the chain to five. By Step~5, Tree-Ring edit-only TPR falls from $0.98$ to $0.061$, while five methods retain $1.00$, including four with decreasing $d'$. Mean TPR over seven subsequent distortions also separates their responses: PRC falls from $0.940$ to $0.870$, whereas HSQR, GS, T2S, and TAG remain near the ceiling. After the first edit, the six methods with informative clean separation retain $0.56$--$0.70$ of their source $d'$; in later steps, higher-separation methods generally retain a larger fraction per edit (\Cref{tab:multistep}; \Cref{fig:app_seq_dist}). These are descriptive trajectories, motivating separation alongside survival rates.

\subsection{Controlled Tests within Methods}
\label{sec:ea6}

\paragraph{Embedding strength: does separation change within a method?}
We vary embedding strength $\alpha$ in HSTR and Tree-Ring under P2P+NTI, InfEdit, and FlowEdit-SD3. \Cref{fig:dprime_intervention}(b) compares clean $d'$ with pre-distortion post-edit $d'$, averaged over five categories. Post-edit separation increases monotonically with $\alpha$ in all six method--editor conditions (Spearman $\rho{=}1.0$; bootstrap $P(\mathrm{monotonic}){\ge}0.9993$). Across 30 aggregates, clean margin alone gives $R^2{=}0.9608$. \Cref{sec:app_ea6_detail} provides the aggregation and resampling details, and \Cref{tab:app_source_quality} documents the accompanying source-fidelity cost on 100 sources per method and strength.

\paragraph{Angular layout: what changes at matched clean separation?}
\label{sec:eb3}
Within HSTR, Compact and Distributed angular supports have equal coefficient budgets and matched radial histograms at clean-$d'$ targets near 5 and 8. Using paired complete sources and their category-mean native scores, \Cref{fig:dprime_intervention}(c) contrasts layout at matched separation with margin targets at fixed layout. All six layout intervals include zero, with point estimates from $-0.144$ to $+0.151$; all six margin intervals exclude zero, with effects from $+1.342$ to $+2.122$. \Cref{sec:app_eb3_extended} reports the controls and full intervals.

These within-method tests target separation; \Cref{tab:app_source_quality,sec:app_eb3_extended} quantify their fidelity costs.

\subsection{Controlling for Severity and Holding Out Methods}
\label{sec:ee2}

We predict \emph{per-image edit-only detection} from clean $d'$ and input-to-output LPIPS across six methods, with repeated source identities (\Cref{sec:app_severity}). On the covered common-LPIPS-support cohort, five-bin reweighting changes Edit-only TPR by at most $0.014$, preserving the observed separation between Tree-Ring and the near-ceiling methods.

Combining $d'$ and LPIPS improves pooled AUC but worsens held-out log-loss for Tree-Ring and the two-method macro, while improving ZoDiac (\Cref{tab:severity_main}). Thus the pooled gain does not transfer reliably. Editor terms add little to pooled AUC (\Cref{sec:app_severity}).

% All values copied from revision 69, Section 5.4 / Appendix E.5.
\begin{table}[!htbp]
\centering
\caption{Per-image edit-only logistic prediction: pooled AUC $\uparrow$ (left) and held-out log-loss $\downarrow$ (right). Macro averages Tree-Ring and ZoDiac; the other four held-out methods are near ceiling. \Cref{sec:app_severity} specifies the models.}
\label{tab:severity_main}
\footnotesize
\begin{minipage}[t]{0.43\linewidth}
\centering
\begin{tabular}{@{}lr@{}}
\toprule
Pooled model & AUC $\uparrow$ \\
\midrule
LPIPS only & 0.633 \\
Clean $d'$ only & 0.916 \\
$d'$ + LPIPS & 0.945 \\
\bottomrule
\end{tabular}
\end{minipage}\hfill
\begin{minipage}[t]{0.55\linewidth}
\centering
\begin{tabular}{@{}lrr@{}}
\toprule
Held-out method & $d'$ only & $+$ LPIPS \\
\midrule
Tree-Ring & 0.648 & 0.845 \\
ZoDiac & 0.089 & 0.072 \\
Macro (two methods) & 0.369 & 0.459 \\
\bottomrule
\end{tabular}
\end{minipage}
\end{table}

\FloatBarrier

\section{Discussion, Limitations, and Conclusion}
\label{sec:conclusion}

\paragraph{Findings and practical use.}
Editing primarily separates Tree-Ring; subsequent distortions expose a broader spectrum organized by clean separation. Sequential edits reveal declining separation beneath ceiling TPR, and within-method interventions connect clean and post-edit separation. Spatial overlap adds little predictive information, while the matched-separation HSTR layout contrast remains unresolved.

\paragraph{Recommended evaluation protocol.}
Report Edit-only survival separately from the mean over subsequent distortions. When TPR approaches its ceiling, include $d'$ where informative score distributions are available. Track separation through at least a short cumulative sequence, such as Object Swap, Style Transfer, and Background Swap (\Cref{fig:teaser}). Check realized FPR on held-out null images and assess threshold recalibration, using cross-fitting to evaluate that procedure (\Cref{sec:app_recalib}).

\paragraph{Limitations.}
The clean-$d'$ relationship is empirical, supported by within-method interventions and first-edit contraction; interventions also change source fidelity (\Cref{tab:app_source_quality,sec:app_eb3_extended}). Mixed held-out transfer makes the diagnostic complementary to direct editing evaluation. Realized FPR varies, but recalibration preserves separated pairs at the main points. Editor settings and distortion severities are not perceptually matched, yet tiers survive all tested weight changes; the Edit-only LPIPS check does not match semantics. GS/TAG restrict continuous-margin comparison but retain native-rule TPR evaluation. RGB spatial tests have qualitative attention-map counterparts. Editing-aware learned watermarks and autoregressive editors such as VAREdit~\citep{mao2026varedit} are outside the evaluated scope; the decomposition applies unchanged, and evaluating these families is a natural next step.

% ===== ===== ===== =====

\subsection*{AI use disclosure}

The research questions and core ideas were developed by the authors. Large language models assisted literature retrieval, drafting and polishing, and research execution: experimental and analysis code, debugging, result checking and interpretation, and figure and table preparation. The authors reviewed the AI-assisted work against experimental evidence and take responsibility for the final code, results, citations, and claims.

\subsection*{Ethics statement}

This work studies the robustness of generative-image watermarks under published diffusion and flow editing methods. Such evaluation can inform both more robust watermark design and attempts at evasion; we report it to clarify existing systems' limits. Experiments use public pretrained models and MS-COCO image-caption data, without recruiting human participants or collecting private personal data.

\subsection*{Reproducibility statement}

The benchmark definitions, operating points, and analysis protocols are documented in \Cref{sec:framework,sec:app_hyperparams,sec:app_external,sec:app_severity}; \Cref{tab:app_editor_settings,tab:app_wm_settings} list the editor and watermark settings. Sample counts, filtering, bootstrap procedures, and validation scope accompany the corresponding analyses. \Cref{sec:app_repro} identifies the core pipeline and the separate result artifacts required for the later controlled analyses. The prompt pairs, per-image scores, null banks, and configurations are planned for a separate release with the code.

\bibliography{references}
\bibliographystyle{preprint}

\clearpage

\appendix
% Keep appendix float pages compact instead of stretching space between figures.
\makeatletter
\setlength{\@fptop}{0pt}
\setlength{\@fpsep}{12pt}
\setlength{\@fpbot}{0pt plus 1fil}
\makeatother

\section{Detection Threshold and False-Positive Analysis}

This appendix details the operating thresholds and false-positive measurements supporting \Cref{sec:benchmark_results}.

\subsection{Null Banks, Operating Rules, and Realized FPR}
\label{sec:app_fpr}

With $n{=}100$ null samples, an empirical $q_{99}$ threshold is estimated near the extreme order statistics and its realized FPR need not equal $1\%$. We therefore hold each original threshold fixed and evaluate it on the disjoint 500-image extension. Condition-averaged FPR is $1.7\%$ for Tree-Ring, $2.0\%$ for ZoDiac, $2.7\%$ for HSTR, $2.8\%$ for HSQR, and $3.4\%$ for PRC. T2S and TAG give $1.0\%$ and $0.8\%$ over pooled key--image decisions. \Cref{tab:app_bank} also reports the corresponding full 600-image measurements. The departures from nominal operation therefore persist after excluding the threshold-setting images.

\paragraph{Bank composition and checks.}
The banks score non-watermarked source images under the corresponding distortion. The original bank uses source indices 0--99 and the extension uses 100--599. Original null and watermarked cohorts share source-caption indices; extension images lie outside the watermarked editing cohort. Threshold selection uses null scores only. T2S and TAG reuse 100 keys across images, so 60,000 key--image scores are not 60,000 independent images. PRC uses a single-key null bank.

The FPR measurements concern source-null images under the specified distortions, not semantically edited non-watermarked outputs.

For the empirical score-based methods, the $q_{99}$ decision rule is strict score $>$ threshold, with linear quantile interpolation. GS instead uses a nominal analytic $1\%$ setting with stored bit-accuracy threshold $147/256$ and an implemented comparison of accuracy $\ge147/256$. The threshold-selection tail and implemented comparison differ by the boundary count: under a fair independent 256-bit model, the implemented tail is approximately $1.0285\%$. This is a model-based reference rather than an empirical GS FPR.

Averaging the 32 method--condition FPR cells for Tree-Ring, ZoDiac, HSTR, and HSQR gives $2.07\%$; including PRC gives a five-method macro average of $2.25\%$. These are descriptive macro averages over repeated method and condition measurements. We do not attach a binomial confidence interval to them because no single false-positive count and independent trial denominator corresponds to either macro. The seven-method recalibration in \Cref{tab:app_recalib} preserves all pairwise orders separated by at least $0.02$ at the main points.

\paragraph{Expanded-bank summary.}
\label{sec:app_bank}

\Cref{tab:app_bank} contrasts the disjoint-extension and full-bank FPR across the eight conditions. The bank composition and key--image observation units are specified in \Cref{sec:app_fpr}.

% Source: E1a fixed calibration-100 threshold evaluation and R7 full-bank reports.
\begin{table}[!htbp]
\centering
\caption{Realized FPR under the original calibration thresholds, averaged over eight conditions. Extension 500 excludes the 100 threshold-setting images; Full 600 includes both cohorts. T2S and TAG pool 100 keys per image, so their key--image decisions are not independent images.}
\label{tab:app_bank}
\begin{tabular}{lrrr}
\toprule
Method & Unit & Extension 500 & Full 600 \\
\midrule
Tree-Ring & image & 0.017 & 0.016 \\
ZoDiac & image & 0.020 & 0.018 \\
HSTR & image & 0.027 & 0.024 \\
HSQR & image & 0.028 & 0.025 \\
PRC & image & 0.034 & 0.030 \\
T2S & key--image & 0.010 & 0.010 \\
TAG & key--image & 0.008 & 0.008 \\
\bottomrule
\end{tabular}
\end{table}

\subsection{Per-Key FPR Decomposition}
\label{sec:app_perkey}

\Cref{tab:app_perkey} reports per-key FPR, overdispersion, and $\chi^2$ tests for T2S and TAG, using 600 images per key. T2S shows no detected between-key heterogeneity within any of the eight conditions at $p{<}0.01$. TAG shows evidence of between-key heterogeneity for Brightness, JPEG, BM3D, and VAE-B. This test concerns variation across keys within a condition; it does not test equality of FPR across conditions. The per-key FPR range is $0$--$2.83\%$.

% Source: iclr2027/R7ext_T2S_bank_report.csv, R7ext_TAG_bank_report.csv
\begin{table}[!htbp]
\centering
\caption{Per-key FPR for T2S and TAG. Overdispersion $>1$ describes variability above the reference level; statistical evidence for key dependence is assessed by the chi-squared $p$ value. $\dagger$ marks $p{<}0.01$.}
\label{tab:app_perkey}
\begin{tabular}{llrrrrr}
\toprule
WM & Cond & Pooled FPR & Min key & Max key & Overdisp. & $p$ \\
\midrule
T2S & Clean & 0.0100 & 0.0033 & 0.0200 & 0.915 & 0.7153 \\
T2S & Brightness & 0.0100 & 0.0017 & 0.0233 & 1.024 & 0.4157 \\
T2S & Contrast & 0.0100 & 0.0033 & 0.0250 & 0.915 & 0.7153 \\
T2S & JPEG & 0.0100 & 0.0017 & 0.0200 & 0.996 & 0.4909 \\
T2S & Blur & 0.0100 & 0.0017 & 0.0233 & 1.061 & 0.3196 \\
T2S & BM3D & 0.0100 & 0.0000 & 0.0233 & 1.153 & 0.1418 \\
T2S & VAE-B & 0.0100 & 0.0000 & 0.0200 & 1.003 & 0.4719 \\
T2S & VAE-C & 0.0100 & 0.0033 & 0.0200 & 1.173 & 0.1146 \\
\midrule
TAG & Clean & 0.0099 & 0.0000 & 0.0283 & 1.143 & 0.1564 \\
TAG & Brightness & 0.0075 & 0.0000 & 0.0233 & 1.522 & 0.0006$\dagger$ \\
TAG & Contrast & 0.0099 & 0.0000 & 0.0217 & 1.184 & 0.1018 \\
TAG & JPEG & 0.0079 & 0.0000 & 0.0217 & 1.386 & 0.0067$\dagger$ \\
TAG & Blur & 0.0075 & 0.0017 & 0.0183 & 1.099 & 0.2343 \\
TAG & BM3D & 0.0076 & 0.0000 & 0.0250 & 1.410 & 0.0045$\dagger$ \\
TAG & VAE-B & 0.0076 & 0.0017 & 0.0217 & 1.578 & 0.0002$\dagger$ \\
TAG & VAE-C & 0.0076 & 0.0000 & 0.0183 & 1.105 & 0.2236 \\
\bottomrule
\end{tabular}
\end{table}

\subsection{Sensitivity to Threshold Choice}
\paragraph{Stricter operating rules.}
\label{sec:app_zerofp}

\Cref{tab:app_zerofp} applies empirical zero-FP thresholds to Tree-Ring, ZoDiac, HSTR, and HSQR. GS instead uses its stricter analytic setting, with a nominal tail target of $10^{-6}$ and the native inclusive comparison. The GS row is not an empirical zero-FP measurement.

% Source: iclr2027/reagg_summary.csv (rule=zero_fp)
\begin{table*}[t]
\centering
\caption{Sensitivity to stricter operating rules (8-condition composite Avg$_5$): empirical zero-FP for four score-based methods and a stricter analytic rule for GS, as specified in the text.}
\label{tab:app_zerofp}
\begin{tabular}{lrrrrr}
\toprule
Method & P2P & P2P+NTI & InfEdit & FE-SD3 & FE-FLUX \\
\midrule
Tree-Ring & 0.25 & 0.36 & 0.32 & 0.29 & 0.32 \\
ZoDiac & 0.75 & 0.82 & 0.86 & 0.83 & 0.83 \\
HSTR & 0.93 & 0.95 & 0.97 & 0.95 & 0.94 \\
HSQR & 0.99 & 0.99 & 1.00 & 0.99 & 0.99 \\
GS & 1.00 & 1.00 & 1.00 & 1.00 & 0.99 \\
\bottomrule
\end{tabular}
\end{table*}

\paragraph{Recalibration on the expanded bank.}
\label{sec:app_recalib}

\Cref{tab:app_recalib} compares original and recalibrated thresholds for the seven empirically thresholded methods using 8-condition composite Avg$_5$ TPR. Here $n$ denotes null-bank images; both columns evaluate the same stored watermarked scores. At the five main points, every pair separated by at least $0.02$ under the original thresholds retains its order: $30/30$ for Edit only and $86/86$ for both Mean7 and Composite8. The maximum absolute composite change is $0.042$. Across all 20 settings, $348$ of $349$ such composite pairs keep their order; the single exception is HSTR versus ZoDiac at the strongest FlowEdit-FLUX setting ($s{=}0.0$).

% Source: recalib_n600_ordering.csv; is_main_point=True, metric=Composite8.
\begin{table*}[t]
\centering
\caption{Seven-method recalibration sensitivity in 8-condition composite Avg$_5$ TPR at the five main operating points: original $q_{99}$ thresholds from 100 calibration images versus thresholds recalibrated on 600 images. $\Delta$ is the unrounded 600-bank minus 100-bank rate.}
\label{tab:app_recalib}
\small
\begin{tabular}{llrrr}
\toprule
Editor & Method & Avg$_5$ ($n{=}100$) & Avg$_5$ ($n{=}600$) & $\Delta$ \\
\midrule
P2P & Tree-Ring & 0.39 & 0.37 & -0.024 \\
 & ZoDiac & 0.84 & 0.83 & -0.012 \\
 & PRC & 0.82 & 0.80 & -0.022 \\
 & HSTR & 0.94 & 0.90 & -0.041 \\
 & HSQR & 0.99 & 0.99 & +0.001 \\
 & TAG & 1.00 & 1.00 & +0.000 \\
 & T2S & 1.00 & 1.00 & +0.000 \\
\addlinespace
P2P+NTI & Tree-Ring & 0.51 & 0.49 & -0.027 \\
 & ZoDiac & 0.89 & 0.88 & -0.005 \\
 & PRC & 0.88 & 0.86 & -0.016 \\
 & HSTR & 0.96 & 0.94 & -0.023 \\
 & HSQR & 1.00 & 1.00 & +0.000 \\
 & TAG & 1.00 & 1.00 & +0.000 \\
 & T2S & 1.00 & 1.00 & +0.000 \\
\addlinespace
InfEdit & Tree-Ring & 0.47 & 0.45 & -0.022 \\
 & ZoDiac & 0.92 & 0.92 & -0.001 \\
 & PRC & 0.94 & 0.93 & -0.009 \\
 & HSTR & 0.97 & 0.95 & -0.022 \\
 & HSQR & 1.00 & 1.00 & +0.001 \\
 & TAG & 1.00 & 1.00 & +0.000 \\
 & T2S & 1.00 & 1.00 & +0.000 \\
\addlinespace
FlowEdit-SD3 & Tree-Ring & 0.44 & 0.42 & -0.023 \\
 & ZoDiac & 0.89 & 0.89 & -0.001 \\
 & PRC & 0.88 & 0.86 & -0.014 \\
 & HSTR & 0.95 & 0.92 & -0.029 \\
 & HSQR & 0.99 & 0.99 & +0.001 \\
 & TAG & 1.00 & 1.00 & +0.000 \\
 & T2S & 1.00 & 1.00 & +0.000 \\
\addlinespace
FlowEdit-FLUX & Tree-Ring & 0.47 & 0.44 & -0.023 \\
 & ZoDiac & 0.89 & 0.89 & +0.001 \\
 & PRC & 0.87 & 0.85 & -0.014 \\
 & HSTR & 0.94 & 0.92 & -0.028 \\
 & HSQR & 0.99 & 0.99 & +0.001 \\
 & TAG & 1.00 & 1.00 & +0.000 \\
 & T2S & 1.00 & 1.00 & +0.000 \\
\addlinespace
\bottomrule
\end{tabular}
\end{table*}

Six-fold source-level cross-fitting trains each threshold on 500 of the 600 null images and evaluates it on the held-out 100. Method-wise condition-averaged FPRs are $0.8$--$1.2\%$, with method--condition values of $0.7$--$1.3\%$. For T2S/TAG, all keys for an image remain in the same fold. These FPRs evaluate the recalibration procedure, not the paper's fixed 100-image thresholds; \Cref{tab:app_bank} retains the latter's realized FPRs.

\FloatBarrier
\section{Extended Benchmark and Clean-Separation Analyses}

This appendix defines the 8-condition composite used by the tabulated strength sweep and clean-$d'$ fits, and groups the supporting evidence into editor settings and categories, edit validity, cross-method fits, and within-method embedding changes. These analyses provide the details supporting \Cref{sec:strength,sec:margin,sec:ea6}.

\subsection{Main-Point 8-Condition Composite}
\label{sec:app_main_legacy8}

\Cref{tab:app_main_legacy8} reports 8-condition composite TPR and the clean $d'$ values used in the diagnostic analyses. Within category $c$, the composite is
\begin{equation}
\mathrm{TPR}^{(8)}_c=\frac{\mathrm{TPR}_{\text{edit-only},c}+\sum_{j=1}^{7}\mathrm{TPR}_{j,c}}{8},
\label{eq:composite8}
\end{equation}
where the seven distortions are applied separately to the same edited output rather than sequentially. The table reports the mean and worst of the five category values. This composite combines the Edit-only and Post-distortion outcomes separated in \Cref{tab:main_verification}, providing the outcome for the tabulated composite sweep and clean-$d'$ fits.

% Values copied from the original main table; legacy aggregates verified against
% audit_reports-astra/_work/ec7_decomposition.csv.
\begin{table}[!htbp]
\centering
\caption{8-condition composite at the main editing operating point. Each cell reports Avg$_5$ / worst over five edit categories after first averaging the Edit-only result and seven separately applied subsequent distortions within each category. This composite is retained for comparison with the full strength sweep and clean-$d'$ analyses; it is distinct from Edit only and Post-distortion mean in \Cref{tab:main_verification}. $^\dagger$TAG's clean $d'$ is finite but uninformative under clean-score saturation; GS's ``---'' denotes $d'$ not computed.}
\label{tab:app_main_legacy8}
\footnotesize
\begin{tabular}{l ccccc r}
\toprule
Method & P2P & P2P+NTI & InfEdit & FE-SD3 & FE-FLUX & Clean $d'$ \\
\midrule
Tree-Ring & 0.39\,/\,0.37 & 0.51\,/\,0.49 & 0.47\,/\,0.40 & 0.44\,/\,0.39 & 0.47\,/\,0.46 & 4.05  \\
ZoDiac    & 0.84\,/\,0.82 & 0.89\,/\,0.87 & 0.92\,/\,0.90 & 0.89\,/\,0.88 & 0.89\,/\,0.89 & 8.03  \\
PRC       & 0.82\,/\,0.79 & 0.88\,/\,0.84 & 0.94\,/\,0.93 & 0.88\,/\,0.87 & 0.87\,/\,0.86 & 9.62  \\
HSTR      & 0.94\,/\,0.93 & 0.96\,/\,0.95 & 0.97\,/\,0.96 & 0.95\,/\,0.94 & 0.94\,/\,0.94 & 10.69 \\
HSQR      & 0.99\,/\,0.99 & 1.00\,/\,0.99 & 1.00\,/\,0.99 & 0.99\,/\,0.99 & 0.99\,/\,0.99 & 20.67 \\
TAG       & 1.00\,/\,1.00 & 1.00\,/\,1.00 & 1.00\,/\,1.00 & 1.00\,/\,1.00 & 1.00\,/\,0.99 & 22.60$^\dagger$ \\
T2S       & 1.00\,/\,0.99 & 1.00\,/\,1.00 & 1.00\,/\,1.00 & 1.00\,/\,1.00 & 1.00\,/\,1.00 & 23.35 \\
GS        & 1.00\,/\,1.00 & 1.00\,/\,1.00 & 1.00\,/\,1.00 & 1.00\,/\,1.00 & 1.00\,/\,1.00 & ---   \\
\bottomrule
\end{tabular}
\end{table}

\paragraph{Sensitivity to composite weights.}
The table below compares nine alternative summaries with Composite8 across all 20 settings. Every alternative preserves the tiers Tree-Ring $<$ \{ZoDiac, PRC\} $<$ HSTR $<$ \{HSQR, TAG, T2S, GS\}; reversals occur only within the near-ceiling group or between ZoDiac and PRC, whose original gaps in those cases are below $0.017$. Doubling the Edit-only weight preserves the full ranking; Mean7 reverses no pair, with one ZoDiac--PRC tie accounting for its $19/20$ exact-order count.

\begin{center}
\small
\setlength{\tabcolsep}{5pt}
\begin{tabular}{lcc}
\toprule
Alternative & Kendall $\tau_b$ range & Non-ceiling order retained / 20 \\
\midrule
Drop Brightness & 0.691--1.000 & 16 \\
Drop Contrast & 0.929--1.000 & 19 \\
Drop JPEG & 0.929--1.000 & 18 \\
Drop Blur & 0.929--1.000 & 18 \\
Drop BM3D & 0.929--1.000 & 18 \\
Drop VAE-B & 0.857--1.000 & 12 \\
Drop VAE-C & 0.929--1.000 & 20 \\
Edit-only weight $\times2$ & 1.000 & 20 \\
Mean7 & 0.982--1.000 & 19 \\
\bottomrule
\end{tabular}
\end{center}
Non-ceiling order refers to Tree-Ring, ZoDiac, PRC, and HSTR. The corresponding clean-$d'$ tiers are $4.05 < \{8.03,9.62\} < 10.69 < \{20.67,22.60,23.35\}$, with TAG's saturation qualification retained in \Cref{tab:app_main_legacy8}.

\subsection{Editor Settings and Category Decomposition}
\label{sec:app_sweep}

\Cref{fig:strength_sweep_main} separates Edit-only TPR and the post-distortion mean across all 160 method--editor--strength cells. \Cref{tab:app_sweep} reports the corresponding 8-condition composite for all 160 cells.

% Source: audit_reports-astra/_work/ec7_decomposition.csv, legacy8_avg5.
\begin{table*}[t]
\centering
\caption{Full sweep of the 8-condition composite Avg$_5$ TPR: eight watermarks, five editors, and four settings (160 cells). Each cell averages the Edit-only TPR and seven separately applied post-edit distortion TPRs within category before averaging five categories. Higher $s$ is closer to reconstruction within an editor, but $s$ is not a common cross-editor severity scale. Bold marks the main-table setting (FlowEdit-FLUX: $s{=}0.2$; all others: $s{=}0.4$).}
\label{tab:app_sweep}
\footnotesize
\setlength{\tabcolsep}{2.6pt}
\begin{tabular}{l rrrr rrrr rrrr}
\toprule
& \multicolumn{4}{c}{P2P} & \multicolumn{4}{c}{P2P+NTI} & \multicolumn{4}{c}{InfEdit} \\
\cmidrule(lr){2-5} \cmidrule(lr){6-9} \cmidrule(lr){10-13}
Method & 0.7 & 0.4 & 0.2 & 0.0 & 0.7 & 0.4 & 0.2 & 0.0 & 0.7 & 0.4 & 0.2 & 0.0 \\
\midrule
Tree-Ring & 0.396 & \textbf{0.391} & 0.381 & 0.384 & 0.516 & \textbf{0.513} & 0.504 & 0.501 & 0.562 & \textbf{0.474} & 0.446 & 0.442 \\
ZoDiac & 0.843 & \textbf{0.841} & 0.831 & 0.823 & 0.893 & \textbf{0.888} & 0.876 & 0.881 & 0.930 & \textbf{0.918} & 0.912 & 0.903 \\
PRC & 0.832 & \textbf{0.824} & 0.815 & 0.806 & 0.889 & \textbf{0.878} & 0.878 & 0.869 & 0.934 & \textbf{0.937} & 0.931 & 0.932 \\
HSTR & 0.939 & \textbf{0.942} & 0.930 & 0.925 & 0.962 & \textbf{0.958} & 0.955 & 0.958 & 0.975 & \textbf{0.970} & 0.970 & 0.971 \\
HSQR & 0.992 & \textbf{0.992} & 0.991 & 0.993 & 0.997 & \textbf{0.997} & 0.996 & 0.994 & 0.997 & \textbf{0.997} & 0.996 & 0.996 \\
TAG & 0.999 & \textbf{1.000} & 0.999 & 0.999 & 0.997 & \textbf{0.998} & 0.998 & 0.998 & 0.998 & \textbf{0.998} & 0.998 & 0.998 \\
T2S & 0.998 & \textbf{0.998} & 0.998 & 0.997 & 0.999 & \textbf{0.999} & 0.998 & 0.997 & 0.998 & \textbf{0.999} & 0.998 & 0.999 \\
GS & 0.999 & \textbf{0.998} & 0.999 & 0.999 & 0.998 & \textbf{0.999} & 0.999 & 0.999 & 0.999 & \textbf{0.999} & 0.999 & 0.999 \\
\bottomrule
\end{tabular}

\vspace{4pt}
\begin{tabular}{l rrrr rrrr}
\toprule
& \multicolumn{4}{c}{FlowEdit-SD3} & \multicolumn{4}{c}{FlowEdit-FLUX} \\
\cmidrule(lr){2-5} \cmidrule(lr){6-9}
Method & 0.7 & 0.4 & 0.2 & 0.0 & 0.7 & 0.4 & 0.2 & 0.0 \\
\midrule
Tree-Ring & 0.582 & \textbf{0.439} & 0.291 & 0.165 & 0.627 & 0.558 & \textbf{0.466} & 0.266 \\
ZoDiac & 0.930 & \textbf{0.890} & 0.787 & 0.587 & 0.930 & 0.914 & \textbf{0.894} & 0.767 \\
PRC & 0.920 & \textbf{0.878} & 0.819 & 0.645 & 0.932 & 0.914 & \textbf{0.867} & 0.675 \\
HSTR & 0.971 & \textbf{0.951} & 0.914 & 0.739 & 0.974 & 0.966 & \textbf{0.945} & 0.794 \\
HSQR & 0.996 & \textbf{0.993} & 0.991 & 0.967 & 0.997 & 0.995 & \textbf{0.992} & 0.957 \\
TAG & 0.998 & \textbf{0.998} & 0.998 & 0.995 & 0.997 & 0.998 & \textbf{0.998} & 0.989 \\
T2S & 0.998 & \textbf{0.999} & 0.998 & 0.990 & 0.998 & 0.998 & \textbf{1.000} & 0.980 \\
GS & 0.998 & \textbf{0.999} & 0.999 & 0.996 & 0.999 & 0.997 & \textbf{0.998} & 0.990 \\
\bottomrule
\end{tabular}
\end{table*}

\paragraph{Category and condition decomposition.}
\label{sec:app_catcond}

Across the 160 cells, Style Transfer is the most frequent worst category, accounting for 77 cells ($48\%$). Object Swap follows at 31 cells ($19\%$), then Attribute/Color Change (22, $14\%$), Background Swap (16, $10\%$), and Pose/Shape Change (14, $9\%$). These counts use the 8-condition composite and describe frequency rather than effect size. Each cell is assigned one category: the first minimum in lexicographically sorted category-name order, before rounding the exported summary. Thirty cells have tied minima; Style Transfer is the unique minimum in 77 cells and belongs to the tied-or-unique minimum set in 91. Thus the reported single-category counts partition the 160 cells rather than count every tie.

\Cref{tab:main_verification} separates Edit only from Post-distortion mean at the main operating point. Tree-Ring's Edit-only Avg$_5$ is $0.632$--$0.804$, compared with $0.986$--$1.000$ for the other methods; Post-distortion mean lowers Tree-Ring to $0.356$--$0.471$ and spreads ZoDiac, PRC, and HSTR between $0.800$ and $0.965$. The seven distortions therefore widen a gap already present after editing and produce more of the intermediate structure represented by the 8-condition composite.

\subsection{Edit Validity and Solver Budget}
\label{sec:app_validity}

\Cref{tab:edit_validity_main} reports the five main operating points. Each row aggregates Tree-Ring, ZoDiac, HSTR, HSQR, and GS over five categories and 100 source captions, giving 2,500 input--output pairs per editor. All metrics use the watermarked editing input as reference. Positive target-alignment deltas indicate average movement toward the requested edit; perceptual distance and structural similarity describe the magnitude and correspondence of that change. These aggregate diagnostics do not certify every pair. \Cref{sec:app_validity_metrics} gives the metric definitions and preprocessing, and \Cref{tab:app_editvalidity} shows the FlowEdit-FLUX budget sweep.

\paragraph{FlowEdit-FLUX solver budget.}
\label{sec:app_flux}

The FlowEdit-FLUX solver budget changes both perceptual distance and target alignment. \Cref{tab:app_editvalidity} reports the tested budgets using the watermarked editing input as reference. The main benchmark uses $n_{\max}{=}23$; the alternatives provide context for that operating point.

% Source: closeout_20260916/locked_B2_all_cells.csv
\begin{table}[!htbp]
\centering
\caption{FlowEdit-FLUX edit validity across solver budgets, relative to the watermarked editing input. Each row aggregates five original methods, five categories, and 100 source captions (2,500 pairs). LPIPS is a distance and DINO a similarity, so their magnitudes indicate change rather than intrinsic improvement. $\Delta$CLIP and $\Delta$BLIP2 ITM report changes in target alignment.}
\label{tab:app_editvalidity}
\begin{tabular}{lrrrr}
\toprule
Budget ($n_\text{max}$) & LPIPS & DINO & $\Delta$CLIP & $\Delta$BLIP2 ITM \\
\midrule
$n_{\max}{=}17$ (nominal) & 0.089 & 0.928 & 0.0031 & -0.024 \\
$n_{\max}{=}23$ (selected) & 0.187 & 0.812 & 0.0177 & 0.064 \\
$n_{\max}{=}28$ & 0.538 & 0.522 & 0.0425 & 0.275 \\
\midrule
SD3 (reference) & 0.205 & 0.751 & 0.0438 & 0.321 \\
\bottomrule
\end{tabular}
\end{table}

\subsection{Global, Condition-Specific, and Editor-Aware Probit Fits}
\label{sec:app_editor_aware}

The cross-method analysis in \Cref{sec:margin} uses method--editor--strength aggregates of 8-condition composite TPR, rather than the per-image binary response of the severity analysis. \Cref{fig:dprime_intervention}(a) displays 30 points at selected settings; the fit and held-out evaluation use the distinct sets specified below.

\paragraph{Global fit.} A global probit link,
\begin{equation}
\mathrm{TPR} = \Phi(a d' + b),
\label{eq:probit}
\end{equation}
is fitted to 20 main-strength cells from Tree-Ring, ZoDiac, HSTR, and HSQR across five editors, with $a{=}0.3161$ and $b{=}{-}1.3834$. On 40 held-out cells from PRC and T2S across five editors and four strengths, its RMSE is $0.086$ (maximum absolute error $0.307$). The PRC/T2S held-out set is only weakly informative about method transfer because T2S lies near the TPR ceiling.

\paragraph{Condition-specific fits and LOMO.}
\label{sec:app_lomo}
A separate procedure fits one link within each of 20 editor--strength conditions. Within each condition, four training points correspond to Tree-Ring, ZoDiac, HSTR, and HSQR, and PRC/T2S provide two held-out points. The corrected implementation minimizes squared CDF residuals using Nelder--Mead, with $\mathrm{ddof}{=}1$ source separation and the unified operating rules. This gives 40 held-out predictions, RMSE $0.053$, and maximum absolute error $0.140$. LOMO instead fits the other five of the six informative methods within each condition and predicts the omitted method. \Cref{tab:app_lomo} reports its method-wise LOMO errors across those conditions. Tree-Ring has the largest mean error, $0.262$, at the low-$d'$ extrapolation end. These errors belong to the condition-specific procedure; \Cref{sec:app_severity} reports the separate held-out logistic analysis.

% Source: iclr2027/P0-4_refit_lomo.csv
\begin{table}[!htbp]
\centering
\caption{Leave-one-method-out evaluation of condition-specific probit fits for 8-condition composite TPR. Mean and maximum absolute prediction errors summarize the 20 editor--strength conditions, in TPR units. These are separate from the global curve and the per-image severity logistic models.}
\label{tab:app_lomo}
\begin{tabular}{lrrrr}
\toprule
Held-out & $d'$ & Position & Mean abs. err & Max abs. err \\
\midrule
Tree-Ring & 4.05 & extrap.(low) & 0.2622 & 0.3329 \\
ZoDiac & 8.03 & interp. & 0.0785 & 0.1923 \\
PRC & 9.62 & interp. & 0.0692 & 0.1395 \\
HSTR & 10.69 & interp. & 0.0120 & 0.0472 \\
HSQR & 20.67 & interp. & 0.0087 & 0.0419 \\
T2S & 23.35 & extrap.(high) & 0.0031 & 0.0204 \\
\bottomrule
\end{tabular}
\end{table}

\begin{samepage}
\paragraph{Editor-aware descriptive fit.} An editor-aware fit uses
\begin{equation}
\mathrm{TPR}_e = \Phi\!\left(\frac{d'-\mu_e}{\sigma_e}\right),
\label{eq:probit_editor}
\end{equation}
where $\mu_e$ and $\sigma_e$ are fitted transition parameters rather than physical erosion statistics. The tested inversion-based editors have smaller fitted transition scales and flatter strength curves; the tested FlowEdit operating points have larger fitted scales and wider TPR variation. These associations summarize the observed cells and are not used as evidence of an editor-family mechanism.

\end{samepage}

\subsection{Controlled Embedding Strength Intervention}
\label{sec:app_ea6_detail}

\begin{figure}[!htbp]
\centering
\includegraphics[width=\linewidth]{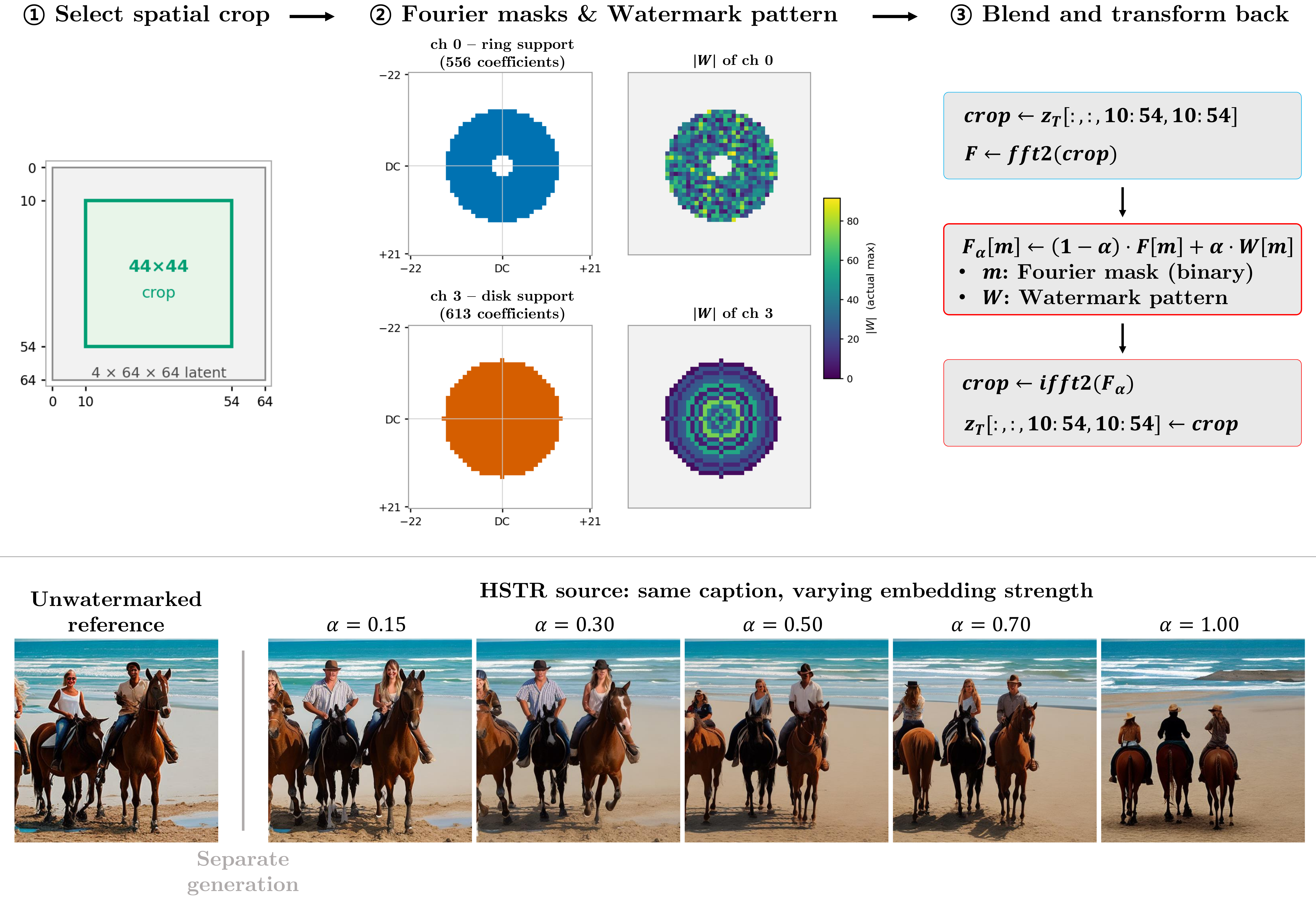}
\caption{HSTR embedding strength intervention. The central $44\times44$ crop of the initial $64\times64$ latent is transformed, and masked coefficients are blended as $F_\alpha[m]=(1-\alpha)F[m]+\alpha W[m]$ before inverse transformation and writeback. The masks and pattern magnitudes are shown separately for channels 0 and 3, using a shared magnitude scale without clipping. The RGB strip shows outputs for caption index 60; the unwatermarked reference is a separate generation with the same caption.}
\label{fig:app_alpha_procedure}
\end{figure}

\paragraph{Design and outcome.}
The strength intervention contains 30 method--editor--$\alpha$ aggregates: HSTR and Tree-Ring, P2P+NTI/InfEdit/FlowEdit-SD3, and $\alpha\in\{0.15,0.30,0.50,0.70,1.00\}$. \Cref{fig:app_alpha_procedure} illustrates blending each original Fourier coefficient with the watermark pattern inside the embedding mask; $\alpha$ is manipulated and $d'$ is measured from the resulting RGB images. The outcome in \Cref{fig:dprime_intervention}(b) is post-edit separation before added distortion: $d'$ is computed separately in each of five categories and then averaged. The stored \texttt{Clean} condition refers to this edited output, not to the pre-edit source. Each category has 100 outputs nominally, with 98--100 common source identities shared across compared strengths, giving 15,000 nominal outputs and 14,989 available. The clean margin reference in \Cref{fig:dprime_intervention}(b) is measured from the watermarked source RGB images at all 100 source indices for each method and strength, using the same detector and source null bank as the existing analysis.

\paragraph{Results and condition-wise extension.}
The plotted 30 aggregates form six method--editor curves before added distortion. A separate condition-wise extension covers all eight detection conditions: monotonic post-edit $d'$ increase holds for both methods and all three editors (48 condition-specific curves). On the 30 plotted aggregates, the clean margin-only model gives $R^2{=}0.9608$; adding method identity changes $R^2$ by $+0.01$ percentage points, and adding editor identity gives a cumulative $+1.88$ points. These are descriptive fits to the 30 aggregates, with repeated source identities across conditions.

\paragraph{Marginal intervals and sensitivity.}
The plotted intervals use 5,000 source bootstrap replicates. Horizontal intervals resample the 100 watermarked source scores while holding the clean null bank fixed. Vertical intervals draw 100 source IDs with replacement from indices 0--99, use the same draw across all five categories, and retain the available IDs within each category. Watermarked and corresponding null scores are resampled together. Each replicate computes category-specific $d'$ and then averages the five values, matching the available-case estimator used for the displayed $y$ point. 

Five InfEdit aggregates have incomplete category coverage; the remaining aggregates have all 100 sources in every category. Complete-case intervals over the common 98--100 IDs are retained as a sensitivity analysis; their endpoints differ from the available-case intervals by at most $0.027$. A separate model sensitivity analysis jointly resamples the 98 IDs common to all conditions; it is distinct from treating the 30 aggregates as independent observations.

\paragraph{Source fidelity.}
\Cref{tab:app_source_quality} measures source fidelity relative to the unwatermarked generation for all 100 source indices per method and strength. LPIPS uses VGG at $512\times512$; PSNR and the script's simple SSIM implementation are included as descriptive means over $n{=}100$ images per row. All three metrics worsen monotonically with $\alpha$, confirming a measured cost to source fidelity; this is not a complete semantic quality--security assessment.

% Source: experiments/45ABCDE_results_received_20260919/2-revision_20260919/
% work_45C_EA_Source_Quality_n100/per_image_quality_n100.csv (1,000 rows).
% The original n=20 data are an exact subset of the combined n=100 file.
\begin{table}[!htbp]
\centering
\caption{Source fidelity across embedding strengths. LPIPS uses VGG at $512\times512$ relative to the unwatermarked generation; SSIM is the source script's simple implementation. Entries are descriptive means over $n{=}100$ images per row.}
\label{tab:app_source_quality}
\small
\begin{tabular}{lrrrr}
\toprule
Method & $\alpha$ & LPIPS & PSNR & SSIM \\
\midrule
HSTR & 0.15 & 0.2827 & 18.19 & 0.6969 \\
 & 0.30 & 0.3847 & 15.89 & 0.5822 \\
 & 0.50 & 0.4499 & 14.55 & 0.5108 \\
 & 0.70 & 0.4834 & 13.67 & 0.4707 \\
 & 1.00 & 0.5170 & 12.47 & 0.4268 \\
\addlinespace
Tree-Ring & 0.15 & 0.2545 & 19.30 & 0.7297 \\
 & 0.30 & 0.3385 & 16.79 & 0.6392 \\
 & 0.50 & 0.4134 & 15.07 & 0.5612 \\
 & 0.70 & 0.4539 & 14.14 & 0.5182 \\
 & 1.00 & 0.5010 & 12.97 & 0.4657 \\
\bottomrule
\end{tabular}
\end{table}

\FloatBarrier
\section{Extended Spatial Analysis}

This appendix provides extended material supporting the spatial analysis in \Cref{sec:spatial}: masking ablation reproducibility, full ablation results for all seven tested watermarks, extended spatial statistics and mean-saliency visualization, and qualitative comparisons of detector sensitivity and editing changes.

\subsection{Masking Ablation Reproducibility}
\label{sec:app_masking_repro}

For the support-matched HSTR/HSQR analysis, the support is the central $352\times352$ pixels, indices $[80,432)$ along each axis of the $512\times512$ RGB image. At $p\in\{5,10,20,30,50\}$ percent, every mask contains $k=\mathrm{round}((p/100)352^2)$ pixels. Targeted (\texttt{high\_support}) and low-saliency (\texttt{low\_support}) masks select the largest and smallest saliency values inside this support using \texttt{argpartition}. Stored mask hashes identify the realized selections.

Random masks (\texttt{random\_support}) sample $k$ support pixels uniformly without replacement, with five draws per image. For source index $i$, percentage $p$, and replication $r\in\{0,\ldots,4\}$, the NumPy generator seed is $20260810+10^6r+10^4i+100p+3$. Selected pixels are replaced by the full source image's channel-wise RGB mean, cast to the image dtype. The score drop is the unmasked minus masked native verification score. The damage ratio divides the mean random drop by the mean targeted drop; the random drop is first averaged over its five draws for each source.

\begin{figure}[!htbp]
\centering
\includegraphics[width=\linewidth]{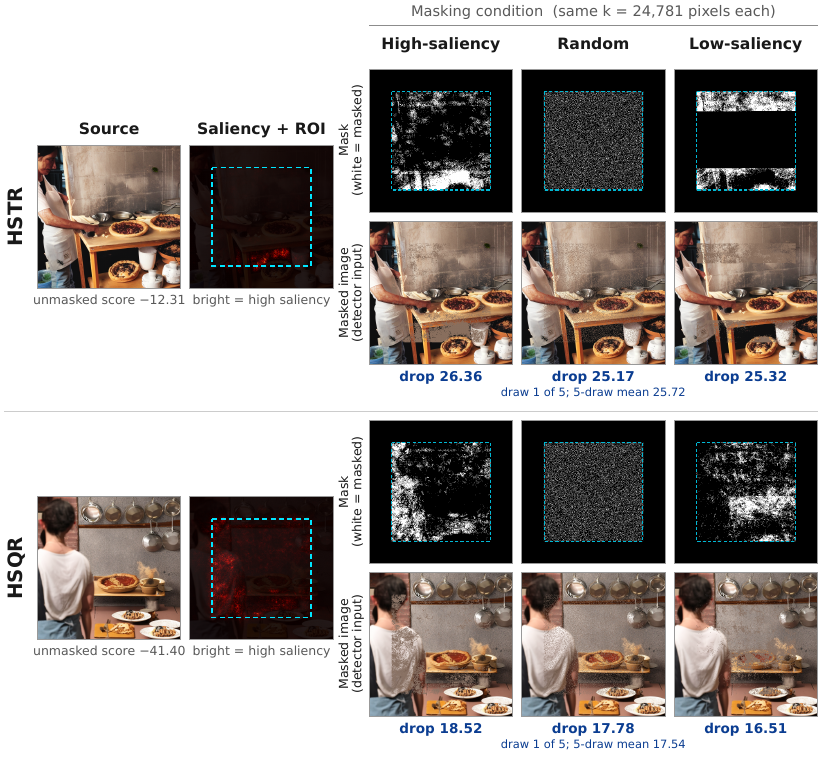}
\caption{Illustrative masking examples for HSTR (top) and HSQR (bottom) on the source for caption index~0, with $p{=}20\%$ of the central $352\times352$ ROI (dashed box), i.e., $k{=}24{,}781$ pixels per mask. Each block shows the source, detection saliency with the ROI, and high-saliency, random (replication~0; the five-draw mean is noted), and low-saliency masks with the resulting detector inputs. White marks masked pixels, which are replaced by the source's channel-wise mean RGB. Displayed drops are stored unmasked-minus-masked native detector scores for these examples, not changes in $d'$ or aggregate random-mask results. These illustrations do not establish targeted--random equivalence or locate the physical watermark signal; \Cref{fig:masking_ablation,tab:app_ablation} give the aggregate comparison.}
\label{fig:app_masking_examples}
\end{figure}

\Cref{fig:app_masking_examples} shows pixels replaced by the source image's channel-wise mean. The single random realization differs from the five-draw aggregate below.

\FloatBarrier
\paragraph{Integrity checks.} The raw file \texttt{ablation\_matched\_v2.csv} contains 7,200 rows with per-trial mask SHA-256 hashes truncated to 16 characters. There are 4,500 unique stored hashes; \texttt{aggregate\_ablation.py} checks this count. This check verifies stored-mask accounting, not independent regeneration of every mask's pixels. Random replications are indexed 0 through 4.

\paragraph{Paired comparison.}
TOST uses the per-source difference between the five-draw random mean and the targeted drop, with equivalence bounds $\pm0.1$ times the mean targeted drop. Equivalence requires both one-sided $t$-test $p$-values below $0.05$. The paired standardized effect $d_z$ uses targeted minus random, with sample standard deviation ($\mathrm{ddof}{=}1$); its sign is therefore opposite to the random-minus-targeted TOST difference. Random draws are averaged within source rather than treated as additional independent images.

\subsection{Masking Ablation Full Results}
\label{sec:app_masking_full}

\begin{figure}[!htbp]
\centering
\includegraphics[width=\linewidth]{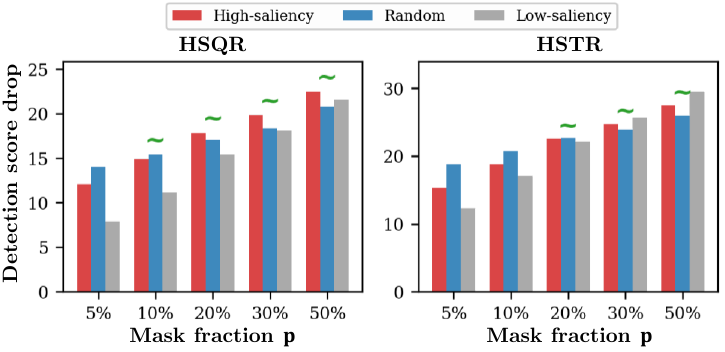}
\caption{Masking ablation on HSQR and HSTR under the support-matched protocol. Targeted (high-saliency) and random masking have similar damage at larger fractions. Low-saliency masking is not uniformly less damaging: HSTR has a larger low- than high-saliency drop at $p{=}30\%$ and $50\%$. Random masking is averaged over five draws per source; $\sim$ marks the stored TOST result with a margin of $\pm10\%$ of the mean targeted drop.}
\label{fig:masking_ablation}
\end{figure}

\Cref{fig:masking_ablation,tab:app_ablation} report the support-matched ablation for HSTR and HSQR, corresponding to the main-text diagnostic in \Cref{sec:ablation}.

% Source: iclr2027/ablation_v2/E4b_summary.csv
\begin{table}[!htbp]
\centering
\caption{Masking ablation: random-to-targeted score-damage ratio and the reported TOST under a 10\% margin. The margin is not independently justified, so the ``Equiv?'' column is a protocol-specific sensitivity result rather than a general equivalence claim.}
\label{tab:app_ablation}
\begin{tabular}{llrrrr}
\toprule
Method & $p$\% & ratio (rnd/high) & $d_z$ & TOST $p$ & Equiv? \\
\midrule
HSQR & 5 & 1.166 & -0.752 & 0.9983 & no \\
 & 10 & 1.036 & -0.211 & 0.0001 & yes \\
 & 20 & 0.953 & 0.353 & 0.0001 & yes \\
 & 30 & 0.925 & 0.655 & 0.0139 & yes \\
 & 50 & 0.924 & 0.787 & 0.0071 & yes \\
\addlinespace
HSTR & 5 & 1.227 & -1.265 & 1.0000 & no \\
 & 10 & 1.101 & -0.747 & 0.5315 & no \\
 & 20 & 1.003 & -0.026 & 0.0000 & yes \\
 & 30 & 0.965 & 0.408 & 0.0000 & yes \\
 & 50 & 0.942 & 0.873 & 0.0000 & yes \\
\addlinespace
\bottomrule
\end{tabular}
\end{table}

\Cref{tab:app_ablation_ext} extends the analysis to five additional watermarks (ZoDiac, PRC, T2S, GS, TAG). Ratios generally move toward one as the mask fraction increases, but convergence is not uniform: GS and TAG remain above one at $p{=}50\%$. The result supports a weak targeted-versus-random distinction under this saliency protocol rather than identical behavior across all seven methods.

% Source: phase4_f2/ablation_ext/ablation_ext_summary.csv
% ratio = mean_drop(random) / mean_drop(high)
\begin{table}[!htbp]
\centering
\caption{Extended masking ablation: random-to-targeted damage ratio for five additional watermarks.
Ratios generally move toward one as the mask fraction increases, but GS and TAG remain above one even at $p{=}50\%$. Their extreme small-$p$ ratios are interpreted as secondary evidence because clean bit accuracy is saturated and saliency uses an STE surrogate (see text).}
\label{tab:app_ablation_ext}
\begin{tabular}{lrrrrr}
\toprule
Method & $p{=}5\%$ & $p{=}10\%$ & $p{=}20\%$ & $p{=}30\%$ & $p{=}50\%$ \\
\midrule
ZoDiac & 1.22 & 1.12 & 1.00 & 0.97 & 0.95 \\
PRC    & 1.23 & 1.10 & 1.02 & 1.00 & 1.00 \\
T2S    & 1.43 & 1.24 & 1.07 & 1.00 & 0.97 \\
\addlinespace
GS     & 80.8 & 20.1 & 4.56 & 2.53 & 1.41 \\
TAG    & 5.60 & 3.11 & 1.91 & 1.45 & 1.20 \\
\bottomrule
\end{tabular}
\end{table}

ZoDiac, PRC, and T2S approach ratios near one by $p{=}20$--$50\%$, broadly resembling the HSTR/HSQR pattern. At $p{=}5\%$, their ratios of $1.22$--$1.43$ indicate more damage from random than targeted masking under the reported ratio. PRC and T2S operate directly in the latent without a Fourier transform, so Fourier domain mixing alone cannot account for the shared descriptive trend; this analysis does not identify an alternative mechanism.

GS and TAG exhibit extreme ratios at small~$p$ (80.8 and 5.60, respectively). Their clean bit-accuracy scores are saturated, and their saliency uses a chosen STE surrogate. We therefore treat these ratios as secondary evidence about the chosen sensitivity ranking. Across the continuous-score methods, ratios occur on both sides of one at larger mask fractions, providing no consistent advantage for targeted masking under this saliency definition.

\subsection{Source Spatial Statistics and Overlap with Editing Changes}
\label{sec:app_spatial_full}

\paragraph{Source-map distribution.} \Cref{tab:app_spatial} pairs clean $d'$ with six spatial statistics. Center ratio, Gini coefficient, and normalized entropy summarize the distribution of the source detection saliency map (\Cref{sec:saliency}) for all eight watermarks, computed over 100 images under each method's support mask. The center ratio is the saliency energy fraction inside the central $352^2$ RGB pixels (area ${\approx}0.47$), distinct from a Fourier embedding-support statistic.

We additionally characterize each map with three complementary statistics: \emph{Moran's~$I$} (larger values indicate stronger positive spatial autocorrelation under the chosen weights), \emph{top-10\% fraction} (energy share of the top decile, $0.1{=}$uniform), and \emph{radial HF/LF ratio} (high- to low-frequency energy of the saliency map itself, ${\approx}0$ indicates low-pass dominance --- this is a spatial-frequency statistic of the image-space saliency map, not the Fourier domain watermark support).

\Cref{tab:app_spatial} reports all six statistics together, and \Cref{fig:mechanism}(a) shows the mean source maps ($n{=}100$ images per method). Its center/global labels describe embedding strategies. Cyan guides indicate the proportional extent of the central $44\times44$ crop of the $64\times64$ initial latent used by HSTR/HSQR before the Fourier transform; the heatmaps describe RGB sensitivity. None of the six metrics shows a monotone ordering aligned with available clean $d'$ or 8-condition composite TPR, consistent with the descriptive observation in \Cref{sec:spatial_disentangle}.

% Source: delocalization/source_stats_summary.csv
\begin{table}[!htbp]
\centering
\caption{Clean separation and six spatial distribution statistics for 8 watermarks (support mask, $n=100$ images). $\dagger$: TAG's finite clean $d'$ is secondary under score saturation; GS has no compatible clean $d'$ estimate. The reported statistics show no monotone ordering aligned with available clean $d^\prime$ or 8-condition composite TPR.}
\label{tab:app_spatial}
\small
\setlength{\tabcolsep}{4pt}
\begin{tabular}{lrrrrrrr}
\toprule
Method & Clean $d'$ & Gini & Entropy$_n$ & Top-10\% & Moran's $I$ & Center & Radial \\
 & & & & frac & & ratio & HF/LF \\
\midrule
Tree-Ring & 4.05 & 0.605 & 0.939 & 0.455 & 0.630 & 0.419 & 0.0026 \\
ZoDiac & 8.03 & 0.605 & 0.940 & 0.454 & 0.659 & 0.402 & 0.0021 \\
PRC & 9.62 & 0.656 & 0.924 & 0.507 & 0.634 & 0.409 & 0.0022 \\
HSTR & 10.69 & 0.662 & 0.919 & 0.514 & 0.622 & 0.901 & 0.0015 \\
HSQR & 20.67 & 0.629 & 0.927 & 0.484 & 0.635 & 0.882 & 0.0014 \\
T2S & 23.35 & 0.680 & 0.919 & 0.526 & 0.645 & 0.431 & 0.0019 \\
TAG & 22.60$^{\dagger}$ & 0.641 & 0.929 & 0.488 & 0.632 & 0.421 & 0.0023 \\
GS & --- & 0.714 & 0.910 & 0.556 & 0.678 & 0.422 & 0.0018 \\
\bottomrule
\end{tabular}
\end{table}

\paragraph{Overlap with image changes and edit-only survival.}
The overlap analysis in \Cref{sec:ec2} uses the RGB pixel-change definition in \Cref{sec:app_edit_region_eqs}. It contains 19{,}963 available image pairs from a nominal 20{,}000 across eight methods, five editors, and five categories: six primary methods with informative continuous margins ($n{=}14{,}971$) and two secondary methods with STE-derived maps (GS, TAG; $n{=}4{,}992$). Mean top-10\% IoU is $0.022$, compared with the independent-uniform geometric reference of $0.053$; both are eight-method descriptive quantities.

Prediction uses edit-only binary survival on the six primary methods. The two logistic models use standardized $d'$ alone or standardized $d'$ plus IoU, with deterministic gradient descent (learning rate $0.05$, 5,000 iterations, L2 coefficient $10^{-4}$); LOMO scaling is fit on training methods only. Defining $\Delta$log-loss as augmented minus $d'$-only, overlap gives $-0.0004$ in-sample and $+0.0005$ under LOMO. Useful non-ceiling held-out evidence is concentrated in Tree-Ring; the other primary methods are at or above approximately $99.3\%$ survival. These image-space diagnostics concern detector sensitivity and do not locate the physical watermark signal.

\subsection{Extended Qualitative Spatial Comparisons}
\label{sec:app_mechanism_ext}

These selected examples complement the aggregate diagnostics in \Cref{fig:mechanism}; they are not a representative sample or a success-rate estimate. Shown $d'_0$ values summarize source banks, and per-image normalization supports comparison of spatial patterns rather than absolute score magnitude.

\begin{figure}[!htbp]
\centering
\includegraphics[width=\linewidth]{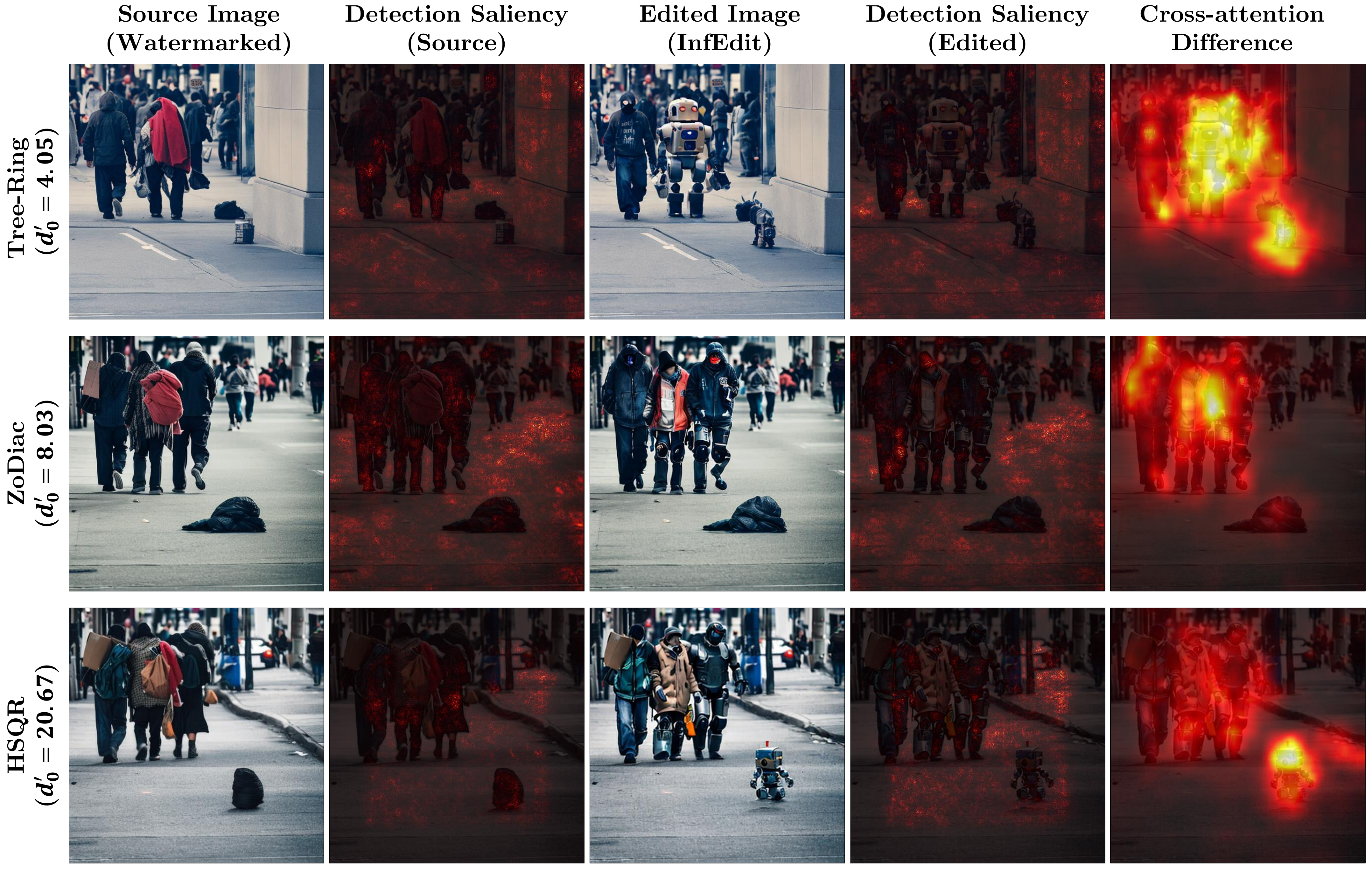}
\caption{Detection saliency before and after an InfEdit Object Swap for Tree-Ring, ZoDiac, and HSQR, alongside cross-attention differences. Rows use method-specific source images with a shared caption index. Each $d'_0$ describes a source bank. Saliency maps visualize detector sensitivity; attention maps summarize changes during editing.}
\label{fig:app_main_spatial_gallery}
\end{figure}

\begin{figure}[!htbp]
\centering
\includegraphics[width=\linewidth]{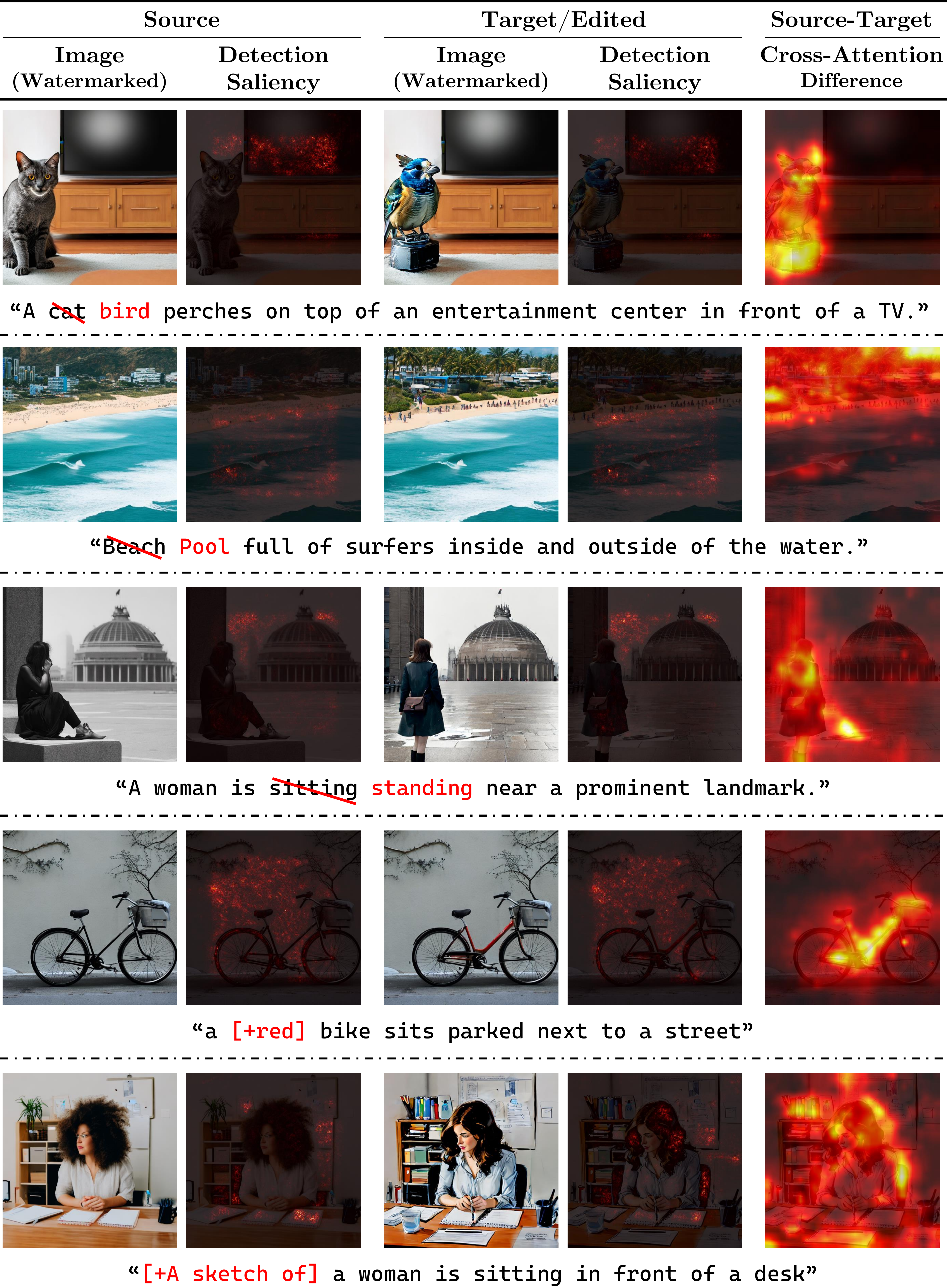}
\caption{HSQR under InfEdit across five semantic edit categories.
Detection saliency is evaluated on the source and edited inputs; cross-attention difference summarizes the editing process.}
\label{fig:app_mechanism_p1}
\end{figure}

\begin{figure}[!htbp]
\centering
\includegraphics[width=0.97\linewidth]{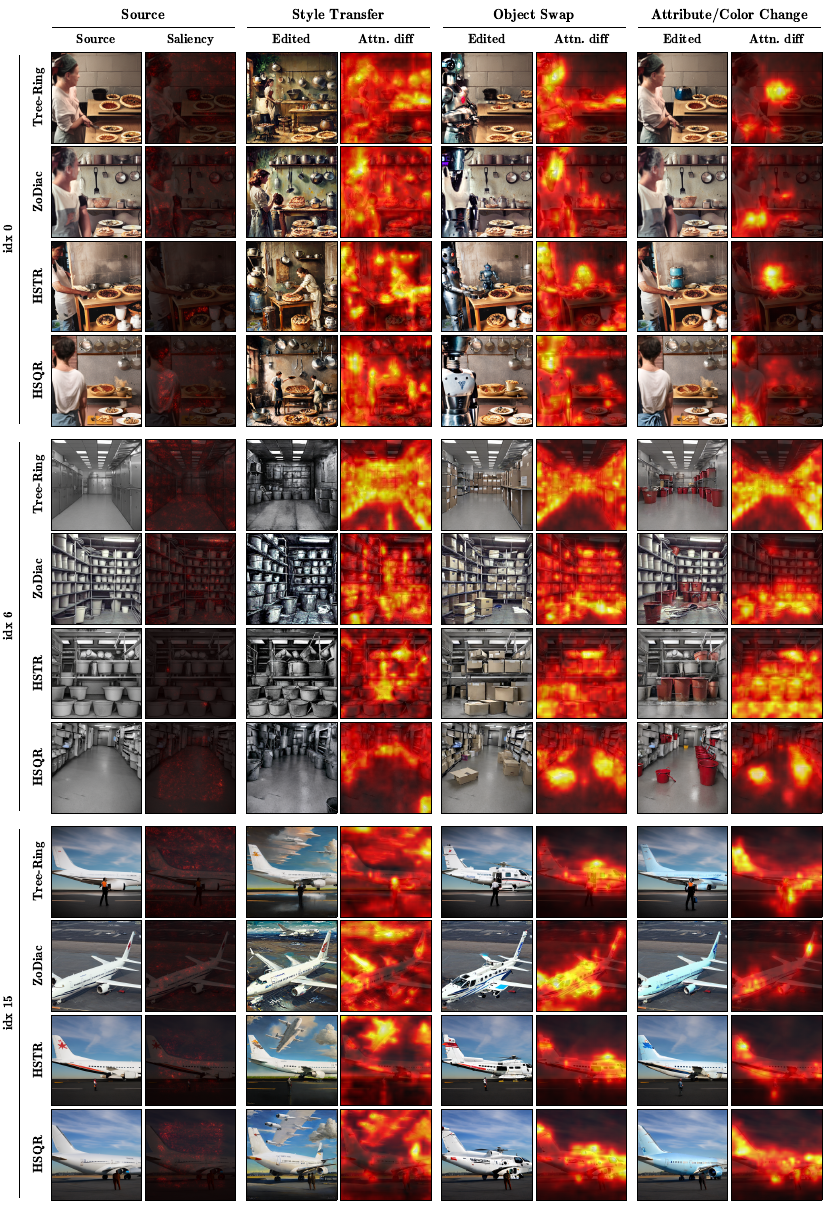}
\caption{Additional InfEdit examples for caption indices~0, 6, and~15 (idx). Rows show Tree-Ring, ZoDiac, HSTR, and HSQR, each with its own watermarked source for the same caption. The first two columns show the source and its detection saliency; each edit category (Style Transfer, Object Swap, Attribute/Color Change) then pairs the edited image with its cross-attention difference. Along a row, categories are compared within one method; down a column, methods are compared under the same edit.}
\label{fig:app_mechanism_p2}
\end{figure}

\paragraph{Source and edited sensitivity.}
\Cref{fig:app_main_spatial_gallery} juxtaposes source and edited saliency with InfEdit attention differences across three methods. \Cref{fig:app_mechanism_p1} extends that before/after comparison to five edit categories for HSQR, while \Cref{fig:app_mechanism_p2} adds three source captions for Tree-Ring, ZoDiac, HSTR, and HSQR, pairing source saliency with InfEdit outputs and cross-attention differences under Style Transfer, Object Swap, and Attribute/Color Change. Together, the figures distinguish detector sensitivity at the input and output from changes in the editing process; attention differences are not RGB change maps, and method-specific source images prevent a pixel-matched comparison between methods.

\begin{figure}[!htbp]
\centering
\includegraphics[width=\linewidth]{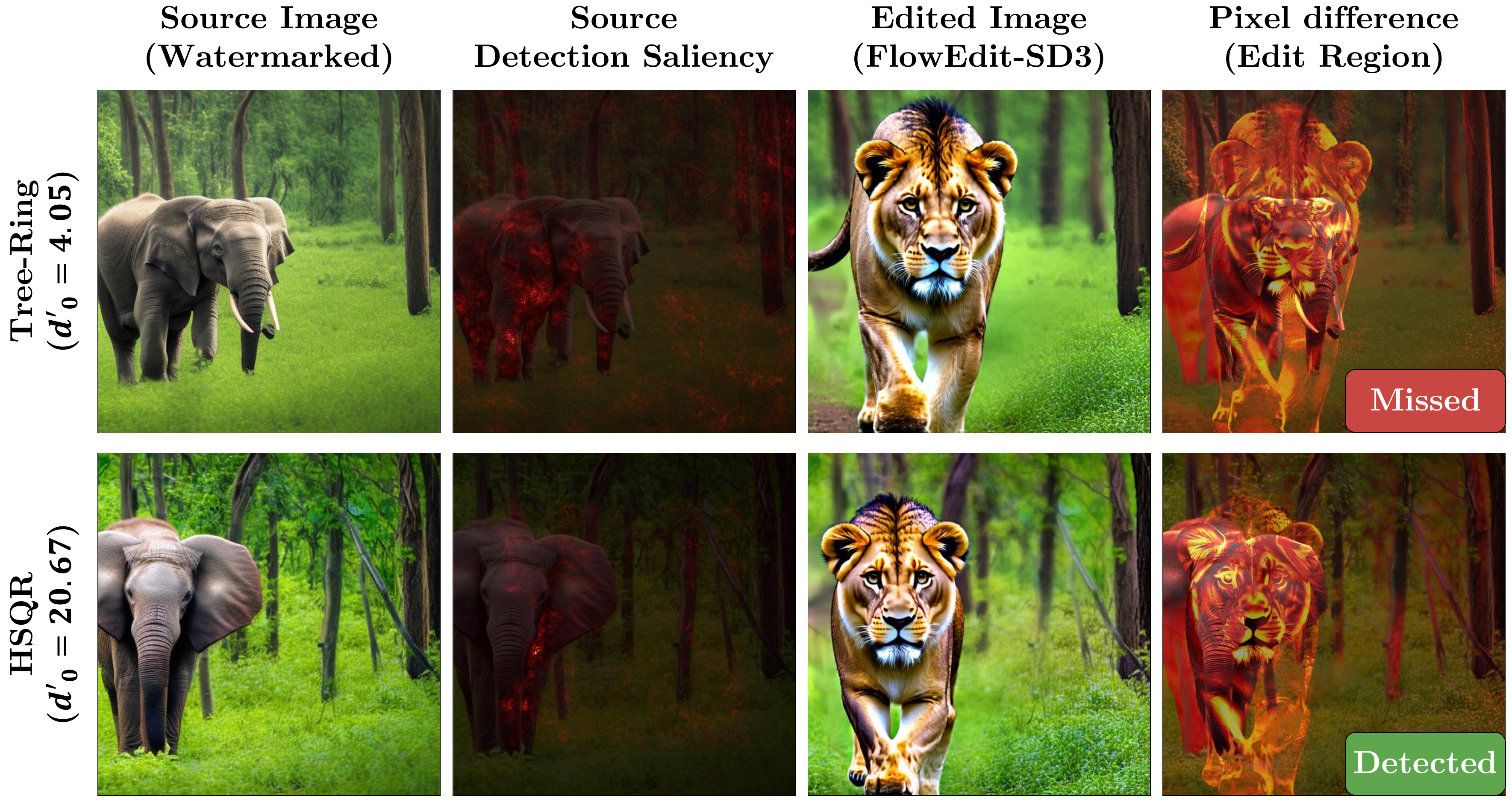}
\caption{Tree-Ring and HSQR under FlowEdit-SD3 ($s{=}0.2$). The final column shows RGB pixel changes between the source and edited image, alongside the verified detection outcome for each displayed case.}
\label{fig:app_mechanism_p3}
\end{figure}

\paragraph{RGB changes under another editor.}
\Cref{fig:app_mechanism_p3} uses FlowEdit-SD3 and directly visualizes source-to-output RGB changes. The displayed Tree-Ring and HSQR cases have different detection outcomes despite both undergoing an object change. This is an illustration of the distinction between visible editing and signal survival, not evidence that spatial overlap causes the difference.

\begin{figure}[!htbp]
\centering
\includegraphics[width=\linewidth]{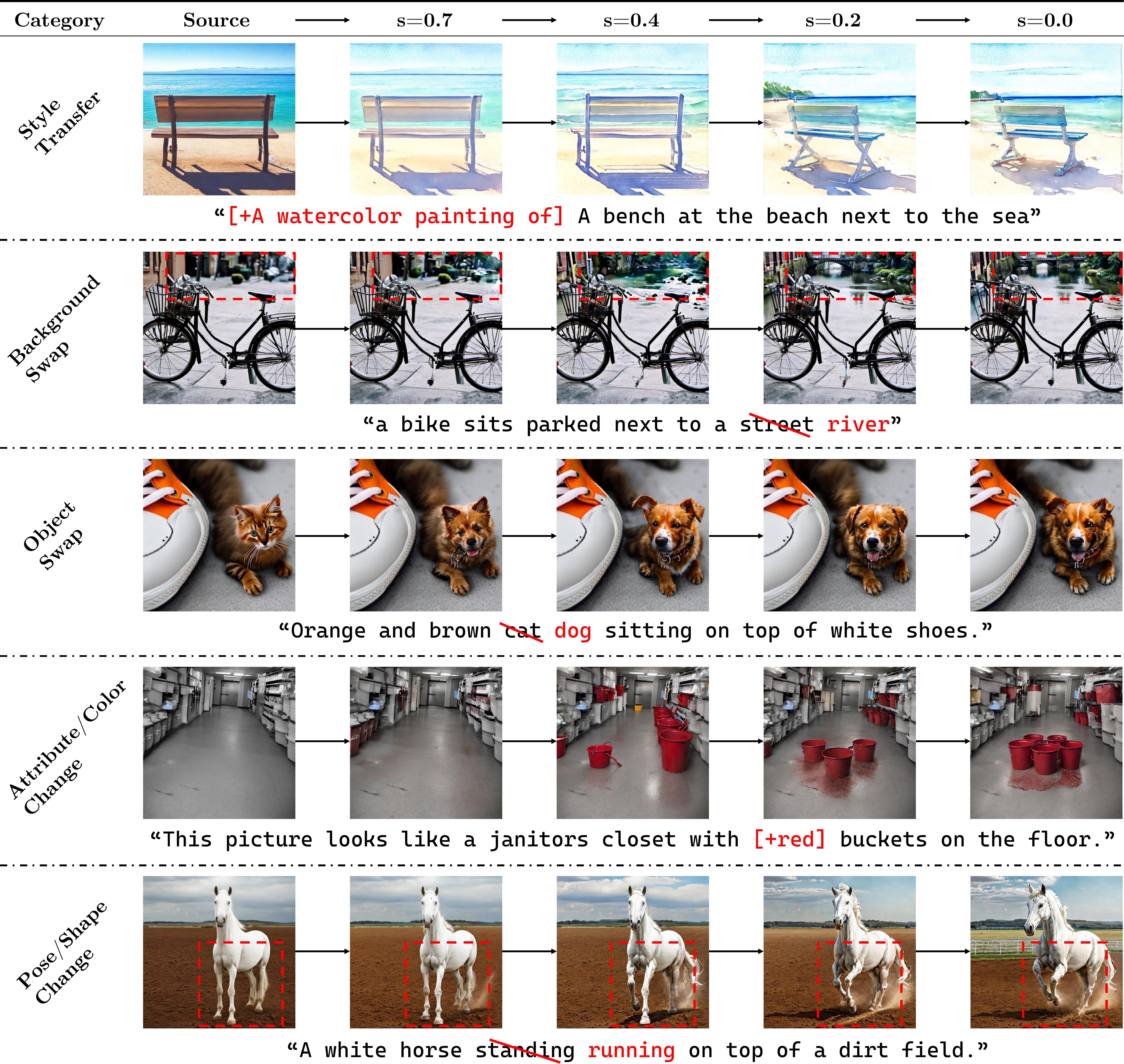}
\caption{HSQR under InfEdit across five editing categories and four editing strengths ($s{=}0.7, 0.4, 0.2, 0.0$).
All twenty edited images maintain successful detection, consistent with HSQR's near-saturated robustness in the main benchmark.}
\label{fig:qualitative}
\end{figure}

\paragraph{Coverage across categories and settings.}
\Cref{fig:qualitative} shows HSQR over five categories and four InfEdit settings. The images put the detected outcomes in visual context across settings; detection in these selected cases neither estimates a population rate nor certifies that every requested edit is successful. Aggregate edit-validity diagnostics are reported in \Cref{sec:app_validity}.

\FloatBarrier

\FloatBarrier
\section{Sequential Editing and Seed Sensitivity}
\label{sec:multistep}

\begin{figure}[!htbp]
\centering
\includegraphics[width=\linewidth,height=0.84\textheight,keepaspectratio]{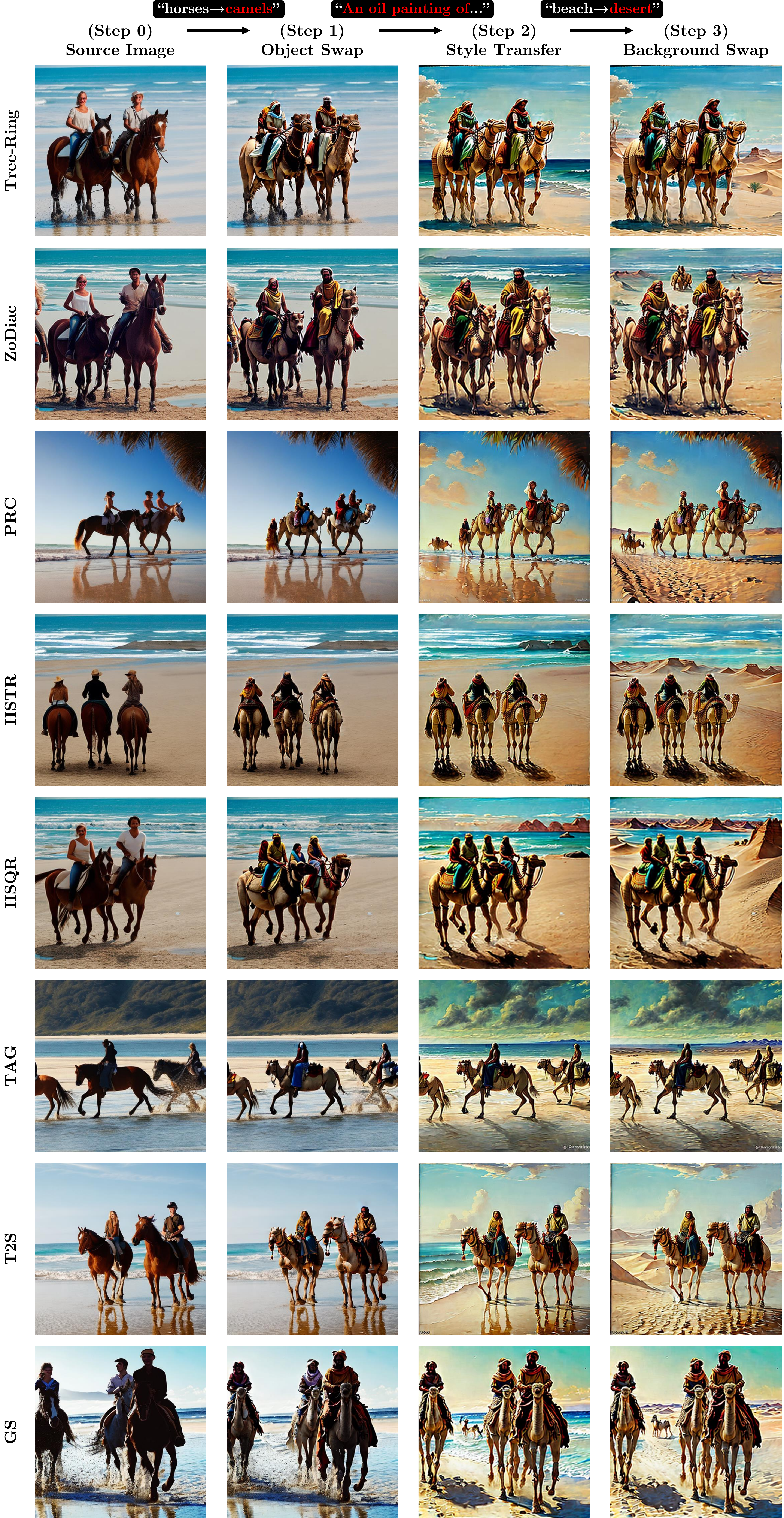}
\caption{Examples of the first three edits in the cumulative InfEdit procedure: Source, Object Swap, Style Transfer, and Background Swap. Every edit takes the preceding output as input. The eight method rows share caption index 60 and use method-specific source images. The displayed prompt changes describe the requested edits; the images show the resulting transformations, including retained background content. The additional fourth and fifth edits are reported numerically in \Cref{tab:multistep,tab:cascade_postdist}.}
\label{fig:app_multistep_examples}
\end{figure}

\Cref{tab:multistep} reports a source baseline followed by the cumulative InfEdit chain Object Swap $\rightarrow$ Style Transfer $\rightarrow$ Background Swap $\rightarrow$ Attribute/Color Change $\rightarrow$ Pose/Shape Change. Each edit takes the preceding output as input. Cross- and self-attention settings are $0.4$, while pipeline denoising strength is $1.0$ with 15 inference steps. TPR and $d'$ are measured before additional distortion; the analysis below applies the seven distortions separately. The empirical $q_{99}$ decision-rule specification and GS analytic exception follow \Cref{sec:app_fpr}. \Cref{fig:app_multistep_examples} illustrates the first three edits for all eight methods.

\paragraph{Coverage.} The source count is 100; Tree-Ring and HSQR use 99 at Steps 1--4, and Step~5 uses 98 for Tree-Ring and 99 for HSQR. Other methods use 100 throughout. Source 94 remains unavailable for Tree-Ring and HSQR after the permitted retry budget of up to 20 new seeds; Tree-Ring additionally loses source 73 at Step~5. These losses are specific to this cascade. TPR is available for all eight methods and $d'$ for seven.

\paragraph{Detection trajectories.} At Step~5, five methods retain TPR $1.00$; four have declining $d'$, while GS has no comparable estimate. The experiment covers one editor and five edits and does not establish eventual failure.

\begin{figure}[!htbp]
\centering
\includegraphics[width=\linewidth,height=0.80\textheight,keepaspectratio]{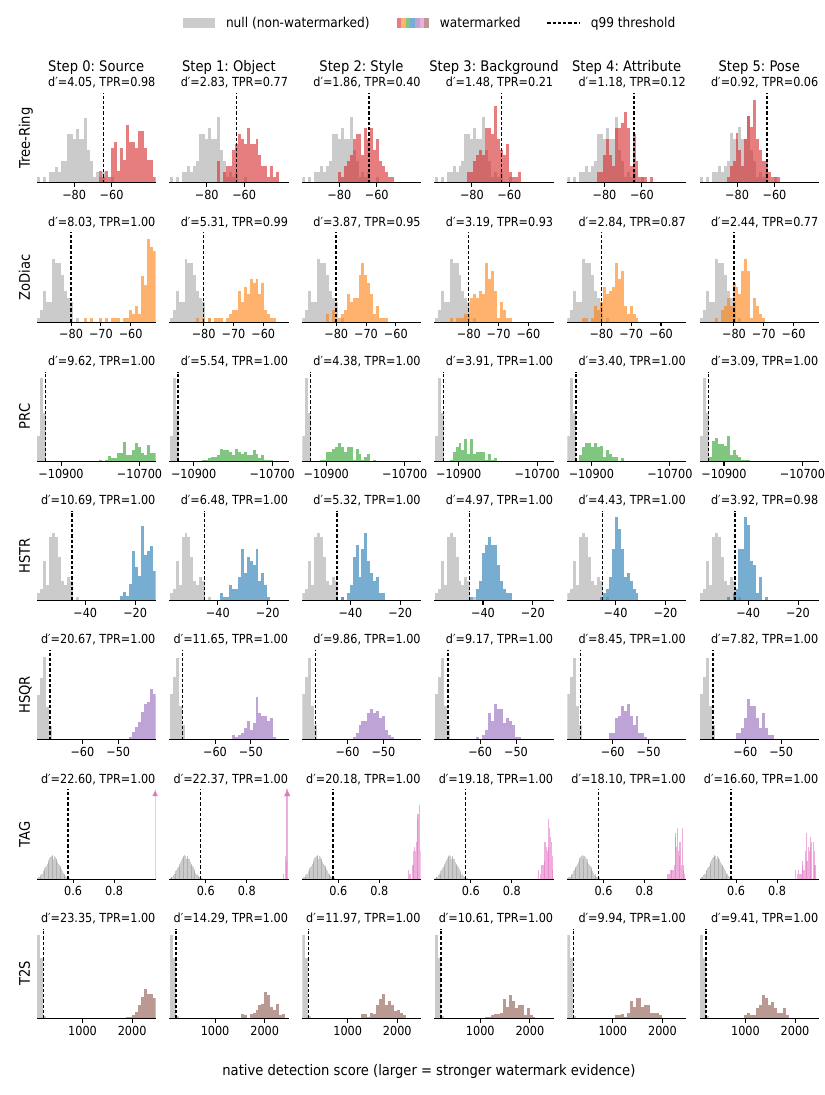}
\caption{Watermarked and null score distributions along the cumulative InfEdit sequence of \Cref{tab:multistep}. Rows show the seven methods with an empirical null; columns show the source (Step~0) and Steps~1--5 (Object Swap, Style Transfer, Background Swap, Attribute/Color Change, Pose/Shape Change) before further distortion. Gray: the unedited, non-watermarked source-null bank that sets each threshold (\Cref{sec:app_fpr}); color: watermarked scores. Scores are in native units, oriented so that larger values indicate stronger evidence, with axes shared within each row. Scores outside the plotted range are omitted: at most six watermarked scores per row, all at Step~0, and under $1\%$ of T2S and TAG null pairs. Dashed lines mark the $q_{99}$ threshold; labels give $d'$ and TPR from \Cref{tab:multistep}. Triangles mark bars truncated at the row's $y$-limit (TAG's source and Step-1 watermarked scores concentrate near bit accuracy $1.0$). GS is omitted because it has no compatible empirical null.}
\label{fig:app_seq_dist}
\end{figure}

\paragraph{Score distributions.} \Cref{fig:app_seq_dist} shows the score distributions behind \Cref{tab:multistep}. Every step is scored against the same unedited null bank, so the visible change along a row is in the watermarked scores. For all seven methods, the mean watermarked score moves toward the null mean at every step, and $d'$ decreases monotonically. Tree-Ring and ZoDiac lose detections as watermarked mass crosses the threshold, whereas PRC, HSQR, T2S, and TAG keep TPR at $1.00$ through Step~5 (HSTR through Step~4) while their separation contracts.

% Source: 45E repaired results/cascade5_full_table_v2.csv
\begin{table}[!htbp]
\centering
\caption{Cumulative InfEdit trajectory: Step 0 is the source; Steps 1--5 add Object Swap, Style Transfer, Background Swap, Attribute/Color Change, and Pose/Shape Change. Each edit takes the preceding output. Cross/self-attention is $0.4$; denoising strength is $1.0$ with 15 inference steps. Values precede further distortion. Source $n{=}100$; Tree-Ring and HSQR use 99 at Steps 1--4, then 98 and 99 at Step 5; all other rows use 100. GS $d'$ is unavailable.}
\label{tab:multistep}
\scriptsize
\setlength{\tabcolsep}{3pt}
\begin{tabular}{l*{12}{r}}
\toprule
 & \multicolumn{2}{c}{Step 0} & \multicolumn{2}{c}{Step 1} & \multicolumn{2}{c}{Step 2} & \multicolumn{2}{c}{Step 3} & \multicolumn{2}{c}{Step 4} & \multicolumn{2}{c}{Step 5} \\
\cmidrule(lr){2-3}\cmidrule(lr){4-5}\cmidrule(lr){6-7}\cmidrule(lr){8-9}\cmidrule(lr){10-11}\cmidrule(lr){12-13}
Method & TPR & $d'$ & TPR & $d'$ & TPR & $d'$ & TPR & $d'$ & TPR & $d'$ & TPR & $d'$ \\
\midrule
Tree-Ring & 0.98 & 4.05 & 0.77 & 2.83 & 0.40 & 1.86 & 0.21 & 1.48 & 0.12 & 1.18 & 0.06 & 0.92 \\
ZoDiac & 1.00 & 8.03 & 0.99 & 5.31 & 0.95 & 3.87 & 0.93 & 3.19 & 0.87 & 2.84 & 0.77 & 2.44 \\
PRC & 1.00 & 9.62 & 1.00 & 5.54 & 1.00 & 4.38 & 1.00 & 3.91 & 1.00 & 3.40 & 1.00 & 3.09 \\
HSTR & 1.00 & 10.69 & 1.00 & 6.48 & 1.00 & 5.32 & 1.00 & 4.97 & 1.00 & 4.43 & 0.98 & 3.92 \\
HSQR & 1.00 & 20.67 & 1.00 & 11.65 & 1.00 & 9.86 & 1.00 & 9.17 & 1.00 & 8.45 & 1.00 & 7.82 \\
TAG & 1.00 & 22.60 & 1.00 & 22.37 & 1.00 & 20.18 & 1.00 & 19.18 & 1.00 & 18.10 & 1.00 & 16.60 \\
T2S & 1.00 & 23.35 & 1.00 & 14.29 & 1.00 & 11.97 & 1.00 & 10.61 & 1.00 & 9.94 & 1.00 & 9.41 \\
GS & 1.00 & -- & 1.00 & -- & 1.00 & -- & 1.00 & -- & 1.00 & -- & 1.00 & -- \\
\bottomrule
\end{tabular}
\end{table}

\subsection{Distortions after Each Cumulative Step}
\label{sec:app_cascade_postdist}

For all eight methods, \Cref{tab:cascade_postdist} reports the undistorted output and Brightness, Contrast, JPEG, Blur, BM3D, VAE-B, and VAE-C applied separately at each step, including the source. Mean 7 averages only the seven subsequent distortions; Comp. 8 averages those and the undistorted output.

The empirical rules compare scores strictly with a condition-specific source-bank $q_{99}$. PRC uses 100 null scores per condition; T2S and TAG use 10,000 key--image scores from 100 keys and 100 null images. The four original continuous-score methods use their corresponding source null scores. Retained watermark and null rows are aligned for Tree-Ring and HSQR: both use 99 at Steps 1--4, then 98 and 99, respectively, at Step~5. Thus the unchanged calibration rule can yield a different numerical threshold as chains become unavailable. GS uses its fixed native $\ge147/256$ rule. All eight conditions score the same cumulative output at a given step; distortions do not create additional editing chains. The undistorted TPR and rounded $d'$ agree with \Cref{tab:multistep}.

From Source to Step~5, Mean~7 falls from $0.664$ to $0.077$ for Tree-Ring, $0.939$ to $0.643$ for ZoDiac, $0.973$ to $0.894$ for HSTR, and $0.940$ to $0.870$ for PRC. PRC's Brightness TPR declines from $0.83$ to $0.52$, illustrating variation hidden by the average. HSQR, GS, T2S, and TAG remain near the ceiling. Some trajectories increase at later steps, so the observations do not establish monotonic degradation for every method. The score-separation trajectories complement these rates without constituting a fitted prediction of Mean~7.

% Source: 45E repaired results/cascade5_full_table_v2.csv; means recomputed before rounding.
\begin{table}[!htbp]
\centering
\caption{Detection after cumulative InfEdit editing, for all eight methods and every subsequent distortion. Each distortion is applied separately to the corresponding cumulative output. Step 0 is the source; Steps 1--5 add Object Swap, Style Transfer, Background Swap, Attribute/Color Change, and Pose/Shape Change. No dist. is the output before further distortion; Mean 7 excludes that column, whereas Comp. 8 includes it. Bright. and Contr. denote brightness and contrast. $n$ is the number of source identities in each row. Empirical $q_{99}$ scores use strict $>$; GS uses native bit accuracy $\ge147/256$.}
\label{tab:cascade_postdist}
\footnotesize
\setlength{\tabcolsep}{2.2pt}
\renewcommand{\arraystretch}{1.0}
\begin{tabular}{lrr*{10}{r}}
\toprule
Method & Step & $n$ & \shortstack{No\\dist.} & Bright. & Contr. & JPEG & Blur & BM3D & VAE-B & VAE-C & \shortstack{Mean\\7} & \shortstack{Comp.\\8} \\
\midrule
Tree-Ring & 0 & 100 & 0.980 & 0.460 & 0.930 & 0.500 & 0.920 & 0.820 & 0.510 & 0.510 & 0.664 & 0.704 \\
 & 1 & 99 & 0.768 & 0.323 & 0.687 & 0.404 & 0.737 & 0.576 & 0.424 & 0.374 & 0.504 & 0.537 \\
 & 2 & 99 & 0.404 & 0.141 & 0.343 & 0.283 & 0.364 & 0.273 & 0.333 & 0.293 & 0.290 & 0.304 \\
 & 3 & 99 & 0.212 & 0.091 & 0.152 & 0.182 & 0.202 & 0.172 & 0.242 & 0.192 & 0.176 & 0.181 \\
 & 4 & 99 & 0.121 & 0.040 & 0.081 & 0.141 & 0.111 & 0.091 & 0.172 & 0.131 & 0.110 & 0.111 \\
 & 5 & 98 & 0.061 & 0.020 & 0.051 & 0.092 & 0.071 & 0.061 & 0.143 & 0.102 & 0.077 & 0.075 \\
\addlinespace[2pt]
ZoDiac & 0 & 100 & 1.000 & 0.690 & 1.000 & 0.960 & 1.000 & 1.000 & 0.990 & 0.930 & 0.939 & 0.946 \\
 & 1 & 100 & 0.990 & 0.650 & 0.970 & 0.940 & 0.980 & 0.970 & 0.970 & 0.950 & 0.919 & 0.927 \\
 & 2 & 100 & 0.950 & 0.400 & 0.950 & 0.870 & 0.940 & 0.890 & 0.860 & 0.880 & 0.827 & 0.843 \\
 & 3 & 100 & 0.930 & 0.290 & 0.890 & 0.840 & 0.880 & 0.810 & 0.930 & 0.890 & 0.790 & 0.807 \\
 & 4 & 100 & 0.870 & 0.220 & 0.800 & 0.760 & 0.770 & 0.740 & 0.910 & 0.860 & 0.723 & 0.741 \\
 & 5 & 100 & 0.770 & 0.140 & 0.690 & 0.690 & 0.640 & 0.630 & 0.920 & 0.790 & 0.643 & 0.659 \\
\addlinespace[2pt]
PRC & 0 & 100 & 1.000 & 0.830 & 1.000 & 0.930 & 1.000 & 1.000 & 0.880 & 0.940 & 0.940 & 0.948 \\
 & 1 & 100 & 1.000 & 0.720 & 1.000 & 0.920 & 1.000 & 1.000 & 0.880 & 0.950 & 0.924 & 0.934 \\
 & 2 & 100 & 1.000 & 0.650 & 1.000 & 0.940 & 1.000 & 0.990 & 0.900 & 0.940 & 0.917 & 0.927 \\
 & 3 & 100 & 1.000 & 0.650 & 1.000 & 0.920 & 1.000 & 1.000 & 0.890 & 0.930 & 0.913 & 0.924 \\
 & 4 & 100 & 1.000 & 0.600 & 1.000 & 0.870 & 1.000 & 0.980 & 0.880 & 0.910 & 0.891 & 0.905 \\
 & 5 & 100 & 1.000 & 0.520 & 1.000 & 0.880 & 1.000 & 0.970 & 0.830 & 0.890 & 0.870 & 0.886 \\
\addlinespace[2pt]
HSTR & 0 & 100 & 1.000 & 0.910 & 1.000 & 0.990 & 1.000 & 0.990 & 0.960 & 0.960 & 0.973 & 0.976 \\
 & 1 & 100 & 1.000 & 0.860 & 1.000 & 0.990 & 1.000 & 0.990 & 0.980 & 0.960 & 0.969 & 0.973 \\
 & 2 & 100 & 1.000 & 0.730 & 1.000 & 1.000 & 1.000 & 0.990 & 0.960 & 0.960 & 0.949 & 0.955 \\
 & 3 & 100 & 1.000 & 0.740 & 1.000 & 1.000 & 1.000 & 0.980 & 0.970 & 0.970 & 0.951 & 0.958 \\
 & 4 & 100 & 1.000 & 0.650 & 0.970 & 0.980 & 1.000 & 0.950 & 0.950 & 0.930 & 0.919 & 0.929 \\
 & 5 & 100 & 0.980 & 0.580 & 0.950 & 0.990 & 0.990 & 0.880 & 0.960 & 0.910 & 0.894 & 0.905 \\
\addlinespace[2pt]
HSQR & 0 & 100 & 1.000 & 0.980 & 1.000 & 1.000 & 1.000 & 1.000 & 1.000 & 1.000 & 0.997 & 0.998 \\
 & 1 & 99 & 1.000 & 0.960 & 1.000 & 1.000 & 1.000 & 1.000 & 1.000 & 1.000 & 0.994 & 0.995 \\
 & 2 & 99 & 1.000 & 0.960 & 1.000 & 1.000 & 1.000 & 1.000 & 1.000 & 1.000 & 0.994 & 0.995 \\
 & 3 & 99 & 1.000 & 0.970 & 1.000 & 1.000 & 1.000 & 1.000 & 1.000 & 1.000 & 0.996 & 0.996 \\
 & 4 & 99 & 1.000 & 0.949 & 1.000 & 1.000 & 1.000 & 1.000 & 0.990 & 1.000 & 0.991 & 0.992 \\
 & 5 & 99 & 1.000 & 0.960 & 1.000 & 1.000 & 1.000 & 1.000 & 1.000 & 1.000 & 0.994 & 0.995 \\
\addlinespace[2pt]
TAG & 0 & 100 & 1.000 & 0.980 & 1.000 & 1.000 & 1.000 & 1.000 & 1.000 & 1.000 & 0.997 & 0.998 \\
 & 1 & 100 & 1.000 & 0.980 & 1.000 & 1.000 & 1.000 & 1.000 & 1.000 & 1.000 & 0.997 & 0.998 \\
 & 2 & 100 & 1.000 & 1.000 & 1.000 & 1.000 & 1.000 & 1.000 & 1.000 & 1.000 & 1.000 & 1.000 \\
 & 3 & 100 & 1.000 & 1.000 & 1.000 & 1.000 & 1.000 & 1.000 & 1.000 & 1.000 & 1.000 & 1.000 \\
 & 4 & 100 & 1.000 & 1.000 & 1.000 & 1.000 & 1.000 & 1.000 & 1.000 & 1.000 & 1.000 & 1.000 \\
 & 5 & 100 & 1.000 & 1.000 & 1.000 & 1.000 & 1.000 & 1.000 & 1.000 & 1.000 & 1.000 & 1.000 \\
\addlinespace[2pt]
T2S & 0 & 100 & 1.000 & 0.990 & 1.000 & 1.000 & 1.000 & 1.000 & 1.000 & 1.000 & 0.999 & 0.999 \\
 & 1 & 100 & 1.000 & 0.990 & 1.000 & 1.000 & 1.000 & 1.000 & 1.000 & 1.000 & 0.999 & 0.999 \\
 & 2 & 100 & 1.000 & 0.990 & 1.000 & 1.000 & 1.000 & 1.000 & 1.000 & 1.000 & 0.999 & 0.999 \\
 & 3 & 100 & 1.000 & 0.990 & 1.000 & 1.000 & 1.000 & 1.000 & 1.000 & 1.000 & 0.999 & 0.999 \\
 & 4 & 100 & 1.000 & 1.000 & 1.000 & 1.000 & 1.000 & 1.000 & 0.990 & 1.000 & 0.999 & 0.999 \\
 & 5 & 100 & 1.000 & 1.000 & 1.000 & 1.000 & 1.000 & 1.000 & 1.000 & 1.000 & 1.000 & 1.000 \\
\addlinespace[2pt]
GS & 0 & 100 & 1.000 & 0.990 & 1.000 & 1.000 & 1.000 & 1.000 & 1.000 & 1.000 & 0.999 & 0.999 \\
 & 1 & 100 & 1.000 & 0.990 & 1.000 & 1.000 & 1.000 & 1.000 & 0.990 & 1.000 & 0.997 & 0.998 \\
 & 2 & 100 & 1.000 & 1.000 & 1.000 & 1.000 & 1.000 & 1.000 & 1.000 & 1.000 & 1.000 & 1.000 \\
 & 3 & 100 & 1.000 & 1.000 & 1.000 & 1.000 & 1.000 & 1.000 & 1.000 & 1.000 & 1.000 & 1.000 \\
 & 4 & 100 & 1.000 & 1.000 & 1.000 & 1.000 & 1.000 & 1.000 & 1.000 & 1.000 & 1.000 & 1.000 \\
 & 5 & 100 & 1.000 & 1.000 & 1.000 & 1.000 & 1.000 & 1.000 & 1.000 & 1.000 & 1.000 & 1.000 \\
\bottomrule
\end{tabular}
\end{table}

\subsection{Historical Sensitivity Check}
\label{sec:app_archive}

An earlier protocol used the same attention setting of $0.4$ but pipeline denoising strength $0.7$, a fixed-generator setup, and a separate zero-FP operating point. Tree-Ring declines from $0.93$ to $0.06$ and ZoDiac from $1.00$ to $0.87$, while HSTR, HSQR, and GS remain at $1.00$. These values are retained as a historical sensitivity result and are not pooled with the current $q_{99}$ cascade. The archive's pipeline denoising value $0.7$ and the current benchmark editor setting $0.4$ are different parameters and are not treated as a direct severity comparison.

\subsection{Separate Seed-Cohort Intervals}
\label{sec:app_seedci}

\Cref{tab:app_seedci} reports a separate repeated-seed cohort for all eight methods across five categories using seed namespaces 1042, 2042, 3042, 4042, and 5042. Source indices are 0--99; retained counts are 95--100 for the original five methods and 96--100 for PRC, T2S, and TAG. The latter contribute 7,417 valid outputs from 7,500 nominal, without retry-based replacement. InfEdit uses cross/self-attention $0.4$ except Pose/Shape Change, which uses cross-attention $0.6$ and self-attention $0.4$. For each method and category, we average TPR over the eight post-edit conditions within each namespace, then report the five-namespace mean and a 95\% Student-$t$ half-width (four degrees of freedom). The intervals summarize variation over editing seeds for the fixed source cohort; PRC, T2S, and TAG intervals use unrounded per-seed values. The maximum half-width remains $0.0444$. These intervals characterize this separate repeated-seed cohort rather than every main benchmark cell.

% Source: iclr2027/seed_ci_v2/seed_ci_full_q99.csv
\begin{table}[!htbp]
\centering
\caption{Verification-TPR intervals from a separate repeated-seed cohort for all eight methods. For each method--category pair, TPR is first averaged over the eight post-edit conditions within each of five disjoint seed namespaces (1042, 2042, 3042, 4042, 5042); the table gives the five-namespace mean and 95\% Student-$t$ half-width (four degrees of freedom). Retained counts are 95--100 for the original five methods and 96--100 for PRC, T2S, and TAG per namespace and category, so these are not intervals for every main benchmark cell.}
\label{tab:app_seedci}
\footnotesize
\begin{tabular}{llrr}
\toprule
Method & Category & Mean TPR & 95\% CI \\
\midrule
Tree-Ring & Attribute\_Color Change & 0.4044 & $\pm$0.0077 \\
 & Background Swap & 0.3931 & $\pm$0.0444 \\
 & Object Swap & 0.3978 & $\pm$0.0216 \\
 & Pose\_Shape Change & 0.4153 & $\pm$0.0071 \\
 & Style Transfer & 0.4137 & $\pm$0.0115 \\
\addlinespace
ZoDiac & Attribute\_Color Change & 0.9000 & $\pm$0.0097 \\
 & Background Swap & 0.8831 & $\pm$0.0110 \\
 & Object Swap & 0.8934 & $\pm$0.0121 \\
 & Pose\_Shape Change & 0.8937 & $\pm$0.0122 \\
 & Style Transfer & 0.8956 & $\pm$0.0127 \\
\addlinespace
HSTR & Attribute\_Color Change & 0.9649 & $\pm$0.0026 \\
 & Background Swap & 0.9734 & $\pm$0.0044 \\
 & Object Swap & 0.9641 & $\pm$0.0075 \\
 & Pose\_Shape Change & 0.9684 & $\pm$0.0048 \\
 & Style Transfer & 0.9667 & $\pm$0.0034 \\
\addlinespace
HSQR & Attribute\_Color Change & 0.9977 & $\pm$0.0032 \\
 & Background Swap & 0.9962 & $\pm$0.0019 \\
 & Object Swap & 0.9962 & $\pm$0.0015 \\
 & Pose\_Shape Change & 0.9959 & $\pm$0.0030 \\
 & Style Transfer & 0.9947 & $\pm$0.0013 \\
\addlinespace
GS & Attribute\_Color Change & 0.9985 & $\pm$0.0013 \\
 & Background Swap & 0.9985 & $\pm$0.0021 \\
 & Object Swap & 0.9995 & $\pm$0.0009 \\
 & Pose\_Shape Change & 0.9997 & $\pm$0.0007 \\
 & Style Transfer & 0.9979 & $\pm$0.0014 \\
\addlinespace
PRC & Attribute\_Color Change & 0.9372 & $\pm$0.0061 \\
 & Background Swap & 0.9360 & $\pm$0.0047 \\
 & Object Swap & 0.9257 & $\pm$0.0102 \\
 & Pose\_Shape Change & 0.9298 & $\pm$0.0076 \\
 & Style Transfer & 0.9300 & $\pm$0.0035 \\
\addlinespace
T2S & Attribute\_Color Change & 0.9990 & $\pm$0.0013 \\
 & Background Swap & 0.9995 & $\pm$0.0009 \\
 & Object Swap & 0.9987 & $\pm$0.0016 \\
 & Pose\_Shape Change & 0.9995 & $\pm$0.0009 \\
 & Style Transfer & 0.9985 & $\pm$0.0013 \\
\addlinespace
TAG & Attribute\_Color Change & 0.9995 & $\pm$0.0014 \\
 & Background Swap & 0.9985 & $\pm$0.0007 \\
 & Object Swap & 0.9990 & $\pm$0.0007 \\
 & Pose\_Shape Change & 0.9985 & $\pm$0.0007 \\
 & Style Transfer & 0.9972 & $\pm$0.0007 \\
\addlinespace
\bottomrule
\end{tabular}
\end{table}

\FloatBarrier
\section{Experimental Protocols}

This appendix documents generation, detection, external integrations, InfEdit filtering, severity adjustment, and the scope of the available reproduction artifacts.

\subsection{Generation and Detection Hyperparameters}
\label{sec:app_hyperparams}

Each watermark uses the default embedding settings of its reference implementation (\Cref{tab:app_wm_settings}); the embedding-strength and layout interventions of \Cref{sec:ea6} are the only changes to embedding. Sources are generated at $512\times512$ with Stable Diffusion v2.1-base~\citep{rombach2022high} (CFG 7.5, 50 DDIM steps, no negative prompt); GS follows the sampler of its official implementation, and ZoDiac embeds post hoc into existing generations of the benchmark captions. \Cref{tab:app_editor_settings} lists the native settings of the five editors, which use SD2.1, LCM DreamShaper v7, SD3~\citep{esser2024scaling}, and FLUX~\citep{blackforestlabs2024flux} backbones; all eight methods are edited with the same implementations and settings. The main point is $s{=}0.4$ except FlowEdit-FLUX, which uses $s{=}0.2$ ($n_{\max}{=}23$).

Each method--editor--strength cell uses 100 source-caption identities and five target categories, giving 500 prompt pairs. Reusing captions across categories and distortions does not increase the number of independent sources. For each semantic edit, Edit only denotes the edited output before further distortion. Post-distortion mean averages seven transformations applied separately afterward: brightness, contrast, JPEG, Gaussian blur, BM3D, and two VAE re-encodings~\citep{balle2018variational,cheng2020learned}. The 8-condition composite averages the Edit-only TPR with those seven TPRs; it is not a sequential seven-distortion cascade. The stored condition name \texttt{Clean} likewise means an edited output without added distortion, whereas ``clean $d'$'' is computed before editing from watermarked-source and null score distributions.

\begin{table}[t]
\centering
\caption{Native editor settings at $s\in\{0.7,0.4,0.2,0.0\}$; bold marks the main operating point. FlowEdit guidance is source/target; InfEdit's Pose/Shape edits use $\tau_{\rm cross}{=}0.6$. All editors seed each source with $42+\mathit{idx}$, shared across categories. P2P and P2P+NTI apply attention replacement or refinement according to the edit type, with local blending on the edited words except in style and pose edits; InfEdit applies attention refinement, with local blending on the target keyword in background, object, and attribute edits.}
\label{tab:app_editor_settings}
\scriptsize
\setlength{\tabcolsep}{3pt}
\begin{tabular}{@{}ll>{\raggedright\arraybackslash}p{0.285\linewidth}llcccc@{}}
\toprule
Editor & Backbone & Procedure & Guidance & Parameter & 0.7 & 0.4 & 0.2 & 0.0 \\
\midrule
P2P & SD2.1-base & DDIM inversion (empty prompt, 50 steps), then 50-step DDIM editing & 7.5 & $\tau_{\rm cross}{=}\tau_{\rm self}$ & 0.7 & \textbf{0.4} & 0.2 & 0.0 \\
P2P+NTI & SD2.1-base & Null-text inversion (50 steps, 10 inner steps), computed once per source; editing as in P2P & 7.5 & $\tau_{\rm cross}{=}\tau_{\rm self}$ & 0.7 & \textbf{0.4} & 0.2 & 0.0 \\
InfEdit & LCM DreamShaper v7 & Inversion-free, 15 LCM steps & 2.0 (source 1.0) & $\tau_{\rm cross}{=}\tau_{\rm self}$ & 0.7 & \textbf{0.4} & 0.2 & 0.0 \\
FlowEdit-SD3 & SD3 Medium & $T{=}50$, $n_{\min}{=}0$ & 3.5 / 13.5 & $n_{\max}$ & 15 & \textbf{30} & 40 & 50 \\
FlowEdit-FLUX & FLUX.1-dev & $T{=}28$, $n_{\min}{=}0$ & 1.5 / 5.5 & $n_{\max}$ & 9 & 17 & \textbf{23} & 28 \\
\bottomrule
\end{tabular}
\end{table}

\paragraph{Sources, prompts, and detection.}
Sources use the first 100 captions of a 5,000-caption MS-COCO 2017 validation subset (one caption per image); the null extension (\Cref{sec:app_fpr}) uses non-watermarked generations for captions 100--599 from the same pipeline. For each caption and category, one target prompt edits the caption's keywords: object, background, and pose terms are substituted, attributes are inserted, and style prefixes such as ``A sketch of'' are added. The 500 prompt pairs follow fixed per-category rules, were checked for rule compliance, and are identical for all eight methods; the sequential edits use a separate five-step prompt chain per caption (\Cref{fig:app_multistep_examples}). All detectors except GS invert queries with DDIM on Stable Diffusion v2.1-base (50 steps, empty prompt, guidance 0); GS uses the inversion of its official implementation. Experiments ran on NVIDIA RTX A6000 and A5000 GPUs, with each watermark and editor in its own software environment.

\paragraph{Distortion implementation.}
The seven post-edit distortions use the following default implementation settings without parameter overrides. Brightness samples a factor rather than multiplying by a constant six. These operators form the fixed stress suite whose weighting sensitivity is assessed in \Cref{sec:app_main_legacy8}.

\begin{center}
\small
\setlength{\tabcolsep}{4pt}
\begin{tabular}{@{}p{0.16\linewidth}p{0.77\linewidth}@{}}
\toprule
Condition & Implementation setting \\
\midrule
Brightness & \texttt{ColorJitter(brightness=6)}; factor sampled uniformly on $[0,7]$ \\
Contrast & PIL contrast enhancement with factor $0.5$ \\
JPEG & PIL JPEG encode/decode, quality $25$; PIL default settings \\
Gaussian blur & OpenCV $5\times5$ kernel, $\sigma=1$ \\
BM3D & RGB denoising with noise scale $0.1$ on $[0,1]$ input \\
VAE-B & CompressAI \texttt{bmshj2018\_hyperprior}, pretrained quality $3$ \\
VAE-C & CompressAI \texttt{cheng2020\_anchor}, pretrained quality $3$ \\
\bottomrule
\end{tabular}
\end{center}

\paragraph{Seeds, retained samples, and averaging.}
The main grid retains one edited output per source--target--editor--setting combination; editor seeds follow \Cref{tab:app_editor_settings}, and InfEdit's recovery pass is recorded separately in \Cref{sec:app_nsfw}. The separate five-namespace study in \Cref{sec:app_seedci} measures seed variation for its fixed source cohort.

TPR uses successfully processed outputs, with no post-hoc quality filter; missing outputs do not count as detection failures. Of the 800 method--editor--strength--category cells, 786 retain all 100 edited outputs and 14 retain 99; all 14 are InfEdit cells. Rates use each category's retained denominator before equal averaging across categories.

\subsection{Edit-Validity Metrics}
\label{sec:app_validity_metrics}

Validity metrics compare every edited output with the watermarked image supplied to the editor. The main validity table aggregates five original methods (Tree-Ring, ZoDiac, HSTR, HSQR, and GS), five categories, and 100 source captions, or 2,500 pairs for each editor or solver budget. LPIPS~\citep{zhang2018unreasonable} uses VGG v0.1 on RGB PIL images resized to $256\times256$ with bicubic interpolation, converted to float32, and scaled as $x/127.5-1$. DINO similarity uses DINOv2-large. $\Delta$CLIP is the input-to-output change in target-text cosine similarity from OpenCLIP~\citep{cherti2023reproducible} ViT-g-14/laion2b\_s12b\_b42k, whereas CLIP-D is the cosine between the image-embedding difference and target-minus-original prompt-embedding difference. $\Delta$BLIP2 ITM is the change in positive image--text matching probability from Salesforce/blip2-itm-vit-g. DINO, CLIP, and BLIP2 retain their model-specific preprocessing and are not asserted to use the LPIPS resizing pipeline. The metrics jointly diagnose change and target alignment but do not guarantee validity for every pair.

\subsection{External Watermark Integration}
\label{sec:app_external}

PRC, T2S, and TAG are integrated from the \texttt{eval\_bench\_wm} repository. Each embeds at generation time through \texttt{WmProvider.get\_wm\_latents()}, followed by the shared DDIM sampling pipeline. The source protocol uses Stable Diffusion v2.1-base, seed $42+\textit{idx}$, 100 MS-COCO prompts, CFG 7.5, and 50 inference steps. The same five editors operate on the resulting pixel-space inputs.

Detection DDIM-inverts the query and computes each native statistic: PRC uses a continuous parity-check likelihood statistic, T2S a recovered-noise $L_1$ statistic, and TAG bit accuracy. PRC uses a single-key null bank; T2S and TAG pool 100 keys over 100 images for calibration. These key--image score counts are distinct from the number of unique images.

\subsection{InfEdit NSFW Protocol}
\label{sec:app_nsfw}

InfEdit uses LCM DreamShaper v7. In a historical asset-recovery pass that included eight strength settings, its safety checker blocked 200 outputs. Retry seeds $1{,}000{,}000+\textit{idx}\times1{,}000+k$ recovered 184; ten benign food images were restored after manual review; six outputs associated with source index 94 were excluded. The six recorded conditions are HSQR Object Swap at $s{=}0.0,0.2,0.7$; HSQR Attribute/Color Change at $s{=}0.7$; and Tree-Ring Attribute/Color Change and Object Swap at $s{=}0.7$. They therefore do not change the main $s{=}0.4$ Table~\ref{tab:main_verification} denominators. The retry ledger is stored in \texttt{infedit\_used\_seeds.csv}; affected analyses use the retained samples. This historical ledger is distinct from the current cascade's two $n{=}99$ chains, the strength intervention's 11 unavailable outputs, and the severity preflight's 129 scattered InfEdit gaps; the latter are not all attributed to the safety checker.

\subsection{Edit Severity: Protocol, Coverage, and Predictive Results}
\label{sec:app_severity}

\paragraph{Coverage and dependence.}
The severity analysis uses 48,893 matched score observations from a nominal grid of 60,000 over six methods, five editors, four strengths, five categories, and 100 outputs per category. Coverage is incomplete; pooled rows repeat source identities and share method-level clean separation values.

\paragraph{Response and features.} The response is the recorded binary native detection decision for each edited output before additional distortion: the \texttt{detected} field in \texttt{ee2\_uniform256.csv}. Thus the logistic models predict individual edit-only decisions rather than aggregate 8-condition composite TPR. Per-image severity is the same LPIPS from watermarked input to edited output defined for validity, matched by source, watermark, editor, strength, and category. It uses the VGG v0.1, RGB PIL $256\times256$ bicubic, float32, $x/127.5-1$ protocol in \Cref{sec:app_validity_metrics}. LPIPS is a severity covariate rather than a guarantee of semantic edit validity. This join contains no DINO, $\Delta$CLIP, or $\Delta$BLIP2 columns. These logistic regressions adjust for LPIPS as a covariate; the separate outcome-blind common-support analysis below restricts and reweights the LPIPS range.

\paragraph{Join accounting.}
The final join contains $48{,}893$ observations from $57{,}871$ preflight records after $9{,}006$ unavailable keys and 28 recovered InfEdit keys; these stages are not a nested exclusion funnel.

\paragraph{Model specification.} Five logistic specifications are compared: LPIPS only, clean $d'$ only, both features in the order $[d',\mathrm{LPIPS}]$, both plus editor terms, and both plus editor interactions. The implementation uses \texttt{StandardScaler} followed by \texttt{LogisticRegression} ($C{=}1$, \texttt{lbfgs}, \texttt{max\_iter}${=}2000$). In-sample results fit the available pooled rows; leave-one-method-out evaluation fits the scaler and classifier on training methods only. Four held-out methods are near the detection ceiling, so the informative macro log-loss summary is limited to Tree-Ring and ZoDiac.

\paragraph{Pooled and held-out results.}
In-sample AUC is $0.633$ for LPIPS only, $0.916$ for clean $d'$ only, and $0.945$ for both. Editor terms and interactions yield $0.949$--$0.950$. For the informative held-out methods, macro log-loss is $0.369$ for $d'$ alone and $0.459$ for $d'$ plus LPIPS. Tree-Ring worsens from $0.648$ to $0.845$, whereas ZoDiac improves from $0.089$ to $0.072$. These logistic results are distinct from the condition-specific probit LOMO errors in \Cref{tab:app_lomo}.

\paragraph{Common-support sensitivity.}
We additionally restrict LPIPS to the outcome-blind intersection $[0.187,0.410]$ of editor--method 5th--95th percentile ranges and form five fixed equal-width bins. Set A covers four methods under all five editors and retains 19,941 of 37,901 observations; Set B covers six methods under P2P, P2P+NTI, and InfEdit and retains 23,389 of 34,893. The adjusted rate is the unweighted mean of the five bin-specific TPRs. \Cref{tab:app_lpips_common_support} shows that this adjustment changes TPR by at most $0.014$ and preserves the method ordering. Strengths and categories are pooled within each method--editor pair; informative variation is concentrated in Tree-Ring, while the other covered methods are at or near ceiling.

% Source: Phase 2 E2_lpips_common_support, outcome-blind five-bin adjustment.
\begin{table}[!htbp]
\centering
\caption{Edit-only TPR on common LPIPS support, pooling all four strengths and five categories. Ranges span P2P, P2P+NTI, InfEdit, FlowEdit-SD3, and FlowEdit-FLUX for Tree-Ring/ZoDiac/HSTR/HSQR, and the first three editors for PRC/T2S. $|\Delta|_{\max}$ is the largest adjusted-minus-unadjusted magnitude on the same retained cohort. Tree-Ring's $0.594$--$0.839$ range differs from \Cref{tab:main_verification}'s $0.632$--$0.804$ because this cohort pools strengths and restricts LPIPS, whereas that table uses the main points.}
\label{tab:app_lpips_common_support}
\small
\setlength{\tabcolsep}{4pt}
\begin{tabular}{lcrrr}
\toprule
Method & Editors & Unadjusted range & Adjusted range & $|\Delta|_{\max}$ \\
\midrule
Tree-Ring & 5 & 0.594--0.839 & 0.582--0.838 & 0.014 \\
ZoDiac & 5 & 0.986--1.000 & 0.985--1.000 & 0.001 \\
PRC & 3 & 0.995--1.000 & 0.993--1.000 & 0.002 \\
HSTR & 5 & 1.000--1.000 & 1.000--1.000 & 0.000 \\
HSQR & 5 & 1.000--1.000 & 1.000--1.000 & 0.000 \\
T2S & 3 & 1.000--1.000 & 1.000--1.000 & 0.000 \\
\bottomrule
\end{tabular}
\end{table}

\subsection{Scope of Reproduction Artifacts}
\label{sec:app_repro}

The earlier core pipeline is recorded at commit \texttt{09e08cf}; \texttt{snapshot\_export.py} exports its processed benchmark results. Watermark and editor implementations use separate environments. Later analyses use additional result artifacts:
\begin{itemize}
\item Embedding strength: \path{ea6_30cell_final.csv}, \path{ci_available_vs_complete_case.csv}, and \path{per_image_quality_n100.csv} for the full source-quality cohort.
\item Angular layout: \path{effects_paired_complete.csv}, with \path{effects_available_category.csv} for sensitivity and \path{source_level_detail.csv} for the paired-source accounting.
\item Sequential editing and seed sensitivity: \path{cascade5_full_table_v2.csv} and the per-namespace result files summarized in \Cref{sec:app_seedci}.
\item Spatial diagnostics: \path{ablation_matched_v2.csv}, \path{source_stats_summary.csv}, and the separately joined saliency--edit overlap records described in \Cref{sec:app_spatial_full}.
\item Severity analysis: \path{ee2_uniform256.csv}, containing the 48,893 matched observations described above.
\end{itemize}
\path{threshold.py} and \path{reaggregate.py} convert the original five methods' native scores to rates; \path{combine_sweep.py} joins external-method exports. Reproduction requires per-image scores and source identities, beyond Edit-only/Mean7/Composite8 aggregates. Masking uses \path{run_ablation_matched.py} and \path{aggregate_ablation.py}; corrected condition-specific probit uses \path{refit_phase5.py}.

Code and supporting artifacts are planned for a separate release.

\FloatBarrier
\section{Score Conventions and Saliency Surrogates}
\label{sec:app_score_functions}

This appendix specifies score conventions and straight-through estimator (STE) surrogates for the saliency map in \Cref{sec:saliency}, followed by definitions of the editing-change maps.

\paragraph{Score orientation and separation.}
Scores are oriented so that larger values indicate stronger watermark evidence. Fourier distance scores use negative distance under this convention; already oriented stored scores retain their sign. Both variances in \Cref{eq:dprime} use Bessel's correction ($\mathrm{ddof}{=}1$). GS has no compatible null estimate for the reported $d'$ analysis, while TAG's clean bit accuracy is saturated.

\paragraph{Image-space sensitivity.}
Saliency is evaluated in the $512\times512$ RGB image space, with DDIM inversion recovering the initial latent where required. These sensitivity maps are distinct from Fourier embedding-support masks. The score function $S(x)$ falls into three families with different gradient computations.

\paragraph{Fourier domain distance.}
Tree-Ring, HSTR, HSQR, and ZoDiac measure the discrepancy between the inverted latent's Fourier coefficients and the ground truth watermark pattern over a predefined frequency mask.
These scores are differentiable through the VAE encoder, DDIM inversion, and FFT, requiring no approximation.

\paragraph{Native continuous statistics.}
PRC computes a parity-check log-likelihood derived from the pseudorandom code structure, while T2S computes an $L_1$ signed-vote norm on the recovered latent.
Both statistics operate directly in the latent domain without a Fourier transform and admit direct backpropagation.

\paragraph{Bit accuracy with STE.}
GS and TAG binarize the latent via a sign decision, $b = \mathrm{sign}(z)$, which blocks gradient flow.
We construct differentiable surrogates using a straight-through estimator (STE): the forward pass computes $\mathrm{sign}(z)$ exactly, while the backward pass uses the derivative of a temperature-scaled sigmoid surrogate $\sigma_T(z) = \mathrm{sigmoid}(z/T)$, i.e.\ $\partial b/\partial z \approx \sigma_T(z)\bigl(1-\sigma_T(z)\bigr)/T$, which is a chosen surrogate gradient for the hard decision.
A key observation is that ChaCha20 decryption in GS, though seemingly nondifferentiable, reduces to an XOR with a fixed keystream at detection time (the key and nonce are known constants). For the XOR description, let $u\in\{0,1\}$ denote the decision in bit coordinates, distinct from the sign-valued $b$ above. With a fixed keystream bit $k$, the mapping is $u$ for $k=0$ and $1-u$ for $k=1$. Its affine extension has derivative $+1$ or $-1$ with respect to $u$. The XOR extension itself introduces no additional approximation to the bit-valued forward operation. In the recovered GS implementation, sign binarization uses the STE with $T=0.1$, and repeated decrypted bits are averaged before computing agreement with the target bits. This mean-vote score is a spatial surrogate, not the native hard-majority decision; a nonzero surrogate gradient does not imply calibrated detection confidence.
This STE-derived surrogate is distinct from each method's native detection statistic and from the informative continuous-margin $d'$ analysis in \Cref{sec:margin}, which is computed for the six methods with native continuous statistics rather than this binarized surrogate.

The native GS decision uses bit accuracy $\ge147/256$. \Cref{sec:app_fpr} details the adjacent analytic threshold-selection tail and the implemented tail probability of approximately $1.0285\%$. This modeled probability is unrelated to the STE used only for spatial analysis.

\subsection{Edit Region Map Definitions}
\label{sec:app_edit_region_eqs}

For P2P, P2P+NTI, and InfEdit, the qualitative attention difference for token $i$ is the normalized absolute difference between source and target cross-attention maps:
\begin{equation}
\Delta A^{(i)} = \mathrm{Norm}\!\left(\frac{1}{|L|\,|T|}\sum_{l \in L,\, t \in T} \bigl|A_{\mathrm{src},l,t}^{(i)} - A_{\mathrm{tgt},l,t}^{(i)}\bigr|\right),
\label{eq:attn_diff}
\end{equation}
where $L$ indexes cross-attention layers and $T$ indexes denoising timesteps. Multi-resolution maps are bilinearly interpolated to $64 \times 64$ before averaging.

For FlowEdit visualizations and for quantitative overlap under every editor, the pixel change map is:
\begin{equation}
\Delta_{\mathrm{pixel}}(x_s, x_e) = \mathrm{Norm}\!\left(\frac{1}{C}\sum_{c=1}^{C}|x_{e,c} - x_{s,c}|\right),
\label{eq:pixel_diff}
\end{equation}
where $x_s$ and $x_e$ are the source and edited RGB images. Attention differences summarize changes in token attention, whereas pixel changes provide the common image-space definition used in the overlap analysis. Neither map is a segmentation annotation of the intended target.

\FloatBarrier
\section{Watermark Method Details}
\label{sec:app_watermark_details}

This appendix provides detailed embedding and detection mechanisms for the eight watermarking methods evaluated in this study, supplementing the summary in \Cref{sec:related}. \Cref{tab:app_wm_settings} lists their settings.

\begin{table}[t]
\centering
\caption{Watermark settings. Each method uses the default embedding settings of its reference implementation; Tree-Ring, HSTR, and HSQR share one Fourier-watermark implementation. Null images are scored with the same keys as watermarked images.}
\label{tab:app_wm_settings}
\scriptsize
\setlength{\tabcolsep}{3pt}
\begin{tabular}{@{}l>{\raggedright\arraybackslash}p{0.47\linewidth}>{\raggedright\arraybackslash}p{0.22\linewidth}l@{}}
\toprule
Method & Embedding & Key / payload & Detection score \\
\midrule
Tree-Ring & Ring pattern in the channel-3 Fourier disk of $z_T$ (radius 14; 613 coefficients) & Per-image key from a 2,048-key codebook & $-L_1$ distance \\
HSTR & Central $44{\times}44$ latent crop: Hermitian-symmetric ring pattern in the channel-3 disk (613 coefficients) and a Gaussian key in the channel-0 ring (556) & Per-image key from a 2,048-key codebook & $-$minimum channel $L_1$ \\
HSQR & QR code (version 1) in channel-3 rFFT coefficients of the central crop (1,764) & Per-image key from a 2,048-key codebook & $-L_1$ distance \\
ZoDiac & Ring pattern in the channel-3 Fourier disk (radius 10; 317 coefficients) of an optimized latent (100 Adam iterations, SSIM target 0.92) & One key & $-L_1$ distance \\
GS & 256-bit message over the $4{\times}64{\times}64$ latent ($8{\times}8$ spatial replication), ChaCha20-encrypted & Per-image key and message & Bit accuracy \\
PRC & Pseudorandom code over 16,384 latent signs & One global key; 512-bit message & Parity log-likelihood \\
T2S & Key in channel 0, message in channels 1--3 & Per-image key; 256-bit message & $L_1$ statistic \\
TAG & 256-bit message repeated across the latent & Per-image message & Bit accuracy \\
\bottomrule
\end{tabular}
\end{table}

\paragraph{Fourier-pattern methods.}
Tree-Ring~\citep{wen2024tree} patterns the Fourier domain ring structure of the initial latent and recovers the watermark via DDIM inversion.
HSTR and HSQR~\citep{lee2025semantic} build on a shared Fourier domain embedding framework.
Their center-aware strategy selects the central $44\times44$ spatial crop, indices $[10{:}54,10{:}54]$ of the $64\times64$ initial latent, applies the Fourier transform to this crop, embeds the pattern in its selected coefficients, and writes the inverse transform back to the crop. The cyan boxes in \Cref{fig:mechanism}(a) indicate this crop proportion. The corresponding saliency maps are measured after image generation and describe detector sensitivity in RGB coordinates.
HSTR recovers a Hermitian symmetric ring similar to the Tree-Ring pattern, while HSQR instead embeds a QR code pattern in the Fourier domain, enabling message encoding with higher capacity.
Unlike the above in-generation methods, ZoDiac~\citep{zhang2025attack} operates as a post-hoc watermark.
Given an existing image, it performs DDIM inversion to obtain the initial noise $z_T$, optimizes the latent to embed a detectable pattern, and regenerates the image.

\paragraph{Continuous-code methods.}
PRC~\citep{gunn2025undetectable} embeds a pseudorandom error-correcting code into the sampling noise and recovers it through soft-decision decoding of the inverted latent.
T2S~\citep{yang2026t2smark} embeds the watermark via Tail-Truncated Sampling and detects it using an $L_1$ norm on the recovered noise.

\paragraph{Bit-encoding methods.}
GS~\citep{yang2024gaussian} encrypts watermark message bits with a ChaCha20 stream cipher and maps each encrypted bit to a Gaussian sample via the inverse CDF.
This procedure preserves the sampling distribution, so the watermarked latent follows the same standard Gaussian distribution as the unwatermarked case.
TAG~\citep{chen2025tag} combines DMJS-based embedding with DVRD, enabling decoding that detects tampering and localizes edited regions.

All eight methods use DDIM inversion to recover the latent-domain signal for detection.

\FloatBarrier
\section{Extended Angular Layout Test at Matched Clean Separation}
\label{sec:app_eb3_extended}

\begin{figure}[!htbp]
\centering
\includegraphics[width=\linewidth]{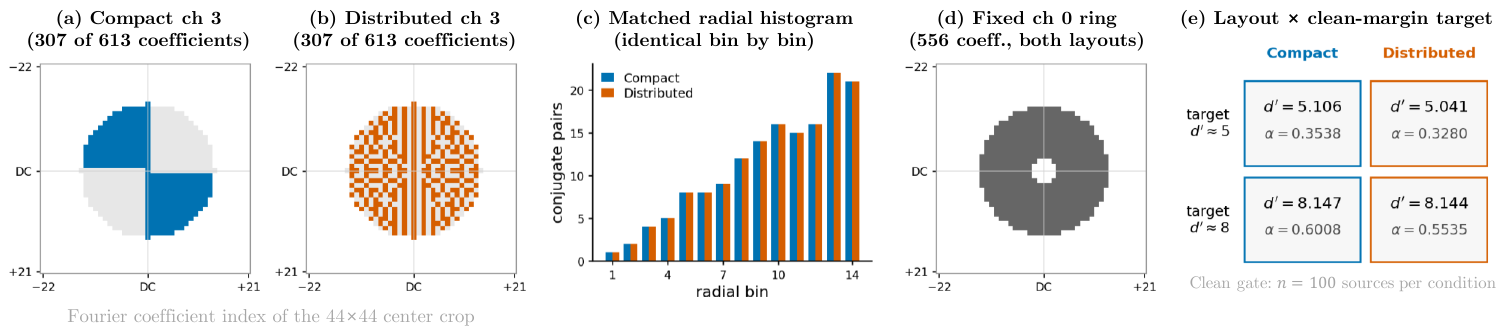}
\caption{HSTR angular layout intervention in the Fourier coordinates of the central $44\times44$ latent crop. Compact and Distributed each select 307 channel 3 coefficients with identical radial pair counts; the 556-coefficient channel 0 ring is fixed. Separate $\alpha$ values achieve clean $d'$ near 5 or 8 on 100 sources per condition. Editing covers 100 sources $\times$ 5 categories $\times$ 3 editors $\times$ 4 conditions: 6,000 nominal and 5,973 available outputs.}
\label{fig:app_layout_procedure}
\end{figure}

This appendix reports the verification of clean separation and the twelve effects computed from scores averaged by source in \Cref{sec:eb3}. \Cref{fig:app_layout_procedure} illustrates the intervention: channel 3 changes its angular support, while channel 0 and the radial histogram are held fixed.

\subsection{Verification of Clean Separation}
\label{sec:app_eb3_gate}

Before editing, the four layout--target conditions were checked on $n{=}100$ source images per condition, not a five-category pool. Their maximum absolute target error is $0.147$, below the recorded tolerance $0.296$; the cross-layout spreads are $0.065$ near $d'{=}5$ and $0.003$ near $d'{=}8$. These values satisfy the clean-separation verification criterion. Source LPIPS uses an AlexNet backbone and the unwatermarked generation as reference.

\begin{center}
\begin{tabular}{lcccc}
\toprule
Condition & Target $d'$ & Measured $d'_{\text{native}}$ & $|\text{error}|$ & Source LPIPS mean \\
\midrule
Compact, $d'{\approx}5$ & 5.0 & 5.106 & 0.106 & 0.3434 \\
Compact, $d'{\approx}8$ & 8.0 & 8.147 & 0.147 & 0.5017 \\
Distributed, $d'{\approx}5$ & 5.0 & 5.041 & 0.041 & 0.3245 \\
Distributed, $d'{\approx}8$ & 8.0 & 8.144 & 0.144 & 0.4959 \\
\bottomrule
\end{tabular}
\end{center}

\subsection{Full Layout and Margin Effects}
\label{sec:app_eb3_effects}

For source $i$, let $\bar{s}_i=(1/5)\sum_c s_{ic}$ be the mean native score over its five edited categories. Each contrast retains only source IDs with all five categories available in both conditions. Each condition's separation substitutes these 94--100 source means and the corresponding layout's 100 null scores into \Cref{eq:dprime}. \Cref{tab:eb3_effects} reports six layout effects (Compact minus Distributed at matched clean $d'$) and six margin effects ($d'{\approx}8$ minus $d'{\approx}5$ at fixed layout), with the paired source count for each contrast. These are differences in separation of scores averaged by source, not differences obtained by pooling edited outputs or averages of five category-specific $d'$ values.

The 95\% intervals use 5,000 percentile-bootstrap iterations. Watermarked source IDs are resampled jointly across compared conditions; the two null score arrays are resampled independently, including when they share a layout label. Thus ``source-aligned'' refers to the edited-source component and does not imply a fully paired or fixed-null bootstrap.

% Source: 45D correction_v2/results/effects_paired_complete.csv
\begin{table}[!htbp]
\centering
\caption{Angular layout and margin effects from paired complete sources (\Cref{sec:app_eb3_effects}). Layout is Compact minus Distributed at matched target $d'$; margin is target $d'{\approx}8$ minus $d'{\approx}5$ at fixed layout. $n$ counts sources with all five categories available in both conditions. The 5,000-iteration percentile bootstrap jointly resamples edited-source IDs and independently resamples the two null arrays. ``0 in CI'' indicates an unresolved effect.}
\label{tab:eb3_effects}
\footnotesize
\setlength{\tabcolsep}{3pt}
\begin{tabular}{llrrrrc}
\toprule
Editor & Comparison & Target $d'$ & $n$ & Effect & 95\% CI & 0 in CI? \\
\midrule
P2P+NTI & Layout & 5 & 100 & $-0.008$ & $[-0.700,\ +0.680]$ & yes \\
P2P+NTI & Layout & 8 & 100 & $-0.144$ & $[-0.996,\ +0.679]$ & yes \\
InfEdit & Layout & 5 & 97 & $+0.083$ & $[-0.620,\ +0.818]$ & yes \\
InfEdit & Layout & 8 & 94 & $+0.097$ & $[-0.820,\ +1.050]$ & yes \\
FlowEdit-SD3 & Layout & 5 & 100 & $+0.096$ & $[-0.506,\ +0.701]$ & yes \\
FlowEdit-SD3 & Layout & 8 & 100 & $+0.151$ & $[-0.499,\ +0.806]$ & yes \\
\addlinespace
P2P+NTI & Margin, Compact & -- & 100 & $+1.793$ & $[+1.044,\ +2.592]$ & no \\
P2P+NTI & Margin, Distributed & -- & 100 & $+1.929$ & $[+1.163,\ +2.759]$ & no \\
InfEdit & Margin, Compact & -- & 97 & $+2.122$ & $[+1.335,\ +3.037]$ & no \\
InfEdit & Margin, Distributed & -- & 95 & $+2.102$ & $[+1.292,\ +2.954]$ & no \\
FlowEdit-SD3 & Margin, Compact & -- & 100 & $+1.397$ & $[+0.755,\ +2.086]$ & no \\
FlowEdit-SD3 & Margin, Distributed & -- & 100 & $+1.342$ & $[+0.702,\ +2.024]$ & no \\
\bottomrule
\end{tabular}
\end{table}

Four layout point estimates are positive and two are negative; every layout interval includes zero. Every margin estimate is positive and every margin interval excludes zero, with effects from $+1.342$ to $+2.122$. As a sensitivity check, averaging each source's available categories retains 100 sources per condition and changes any point estimate by at most $0.015$; the interval conclusions are unchanged. The primary analysis retains the fixed five-category mean. Source LPIPS also increases at the higher margin target in both layouts, so the margin contrast is not an intervention that preserves source fidelity.

\paragraph{Scope.}
\label{sec:app_eb3_scope}

The design uses source indices 0--99, five categories, and 12 editor--layout--target combinations. P2P+NTI and FlowEdit-SD3 each retain all 2,000 outputs; InfEdit retains 1,973, with 27 safety-filtered outputs unavailable. Paired complete-source counts are 100 for the former editors and 94--97 for InfEdit. The scope is HSTR, Compact versus Distributed at 307 channel 3 coefficients, two clean margin targets, and three editors. Layout intervals allow effects of either sign and do not establish equivalence.

\FloatBarrier

\end{document}